\documentclass[10pt,twocolumn,letterpaper]{article}

\usepackage{iccv}
\usepackage{times}
\usepackage{epsfig}
\usepackage{graphicx}
\usepackage{amsmath}
\usepackage{amssymb}

\usepackage[pagebackref=true,breaklinks=true,letterpaper=true,colorlinks,bookmarks=false]{hyperref}

\iccvfinalcopy 

\def\iccvPaperID{146} 

\ificcvfinal\fi
\begin{document}

\title{Learning Uncertain Convolutional Features for Accurate Saliency Detection}

\author{Pingping Zhang\quad Dong Wang\quad Huchuan Lu\thanks{Prof.Lu is the corresponding author.}\quad Hongyu Wang\quad Baocai Yin\\
Dalian University of Technology, China\\
{\tt\small jssxzhpp@mail.dlut.edu.cn \{wdice,lhchuan,whyu,ybc\}@dlut.edu.cn}
}

\maketitle
\thispagestyle{empty}
\begin{abstract}

Deep convolutional neural networks (CNNs) have delivered superior performance in many computer vision tasks.
In this paper, we propose a novel deep fully convolutional network model for accurate salient object detection.
The key contribution of this work is to learn deep uncertain convolutional features (UCF), which encourage the robustness and accuracy of saliency detection.
We achieve this via introducing a reformulated dropout (R-dropout) after specific convolutional layers to construct an uncertain ensemble of internal feature units.
In addition, we propose an effective hybrid upsampling method to reduce the checkerboard artifacts of deconvolution operators in our decoder network.
The proposed methods can also be applied to other deep convolutional networks.
Compared with existing saliency detection methods, the proposed UCF model is able to incorporate uncertainties for more accurate object boundary inference.
Extensive experiments demonstrate that our proposed saliency model performs favorably against state-of-the-art approaches.
The uncertain feature learning mechanism as well as the upsampling method can significantly improve performance on other pixel-wise vision tasks.
\end{abstract}

\section{Introduction}
\begin{figure}
\centering
\begin{tabular}{@{}c}
\includegraphics[width=0.85\linewidth,height=2cm]{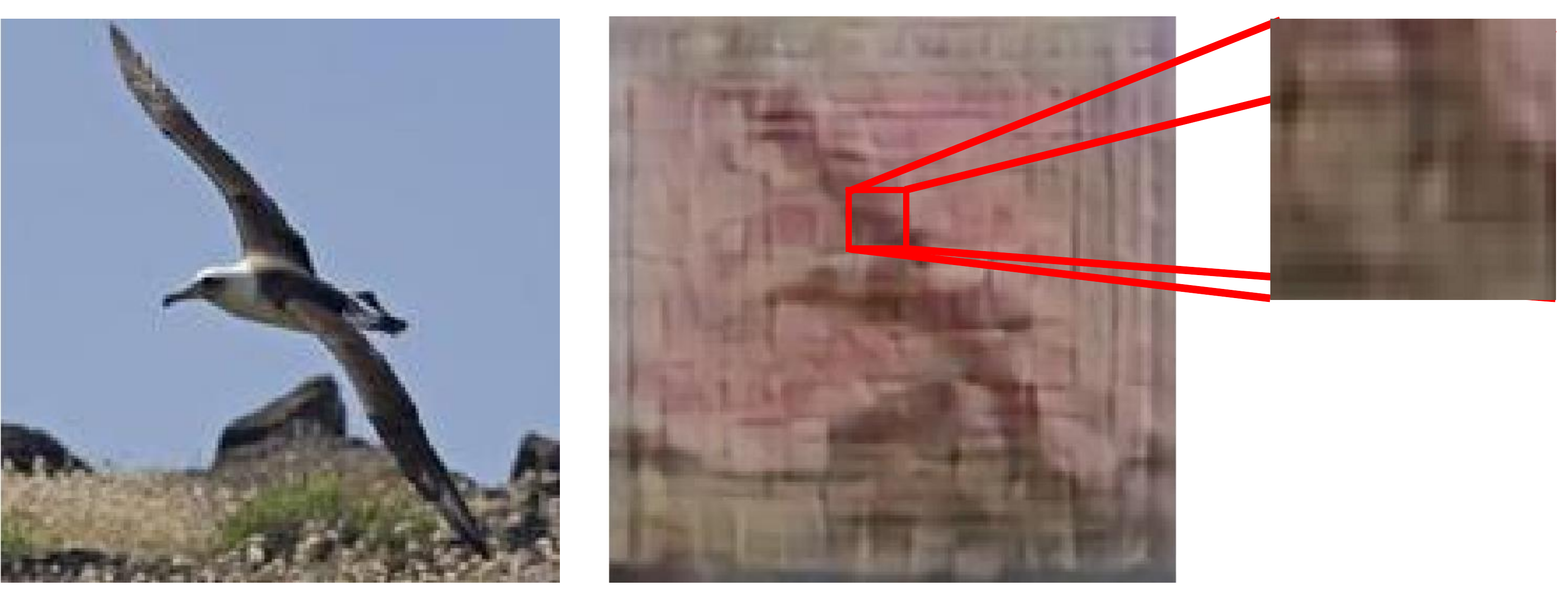}\ \\
{\small(a) Feature Visualization~\cite{dosovitskiy2015inverting}}\ \\
\includegraphics[width=0.85\linewidth,height=2cm]{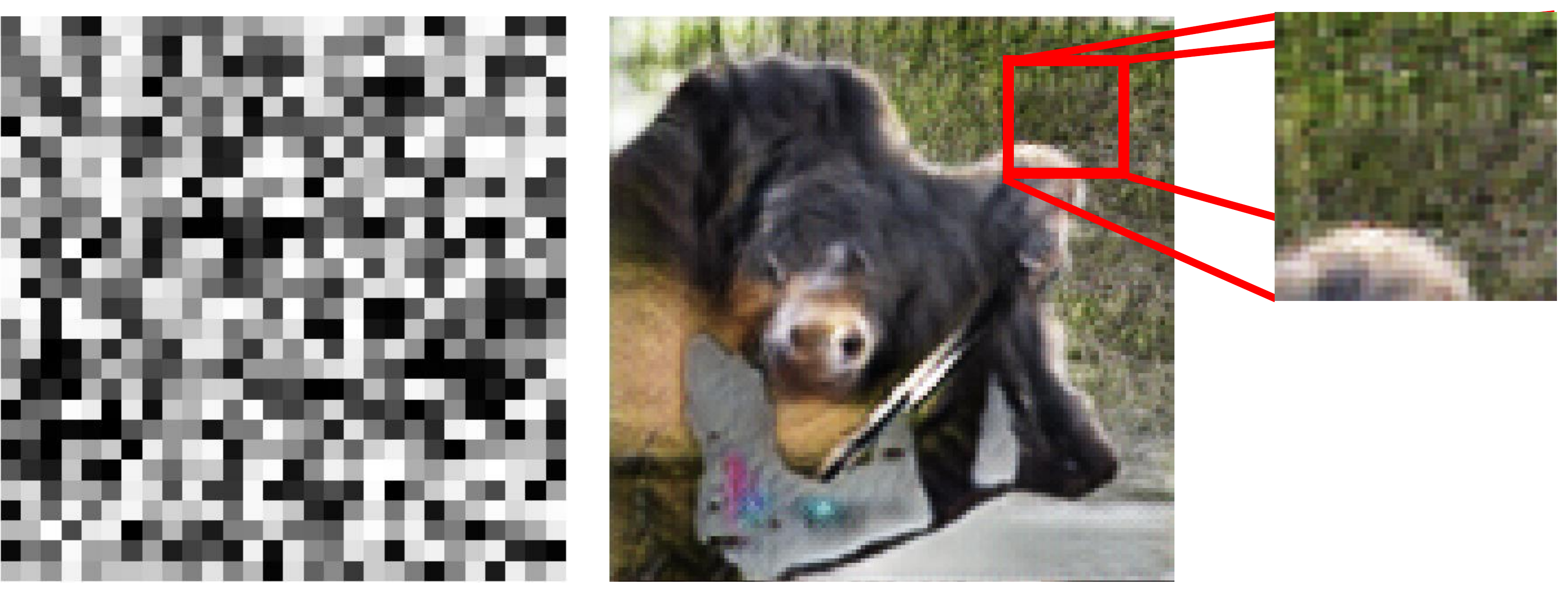}\ \\
{\small(b) Generative Adversarial Example~\cite{salimans2016improved}}\ \\
\includegraphics[width=0.85\linewidth,height=2cm]{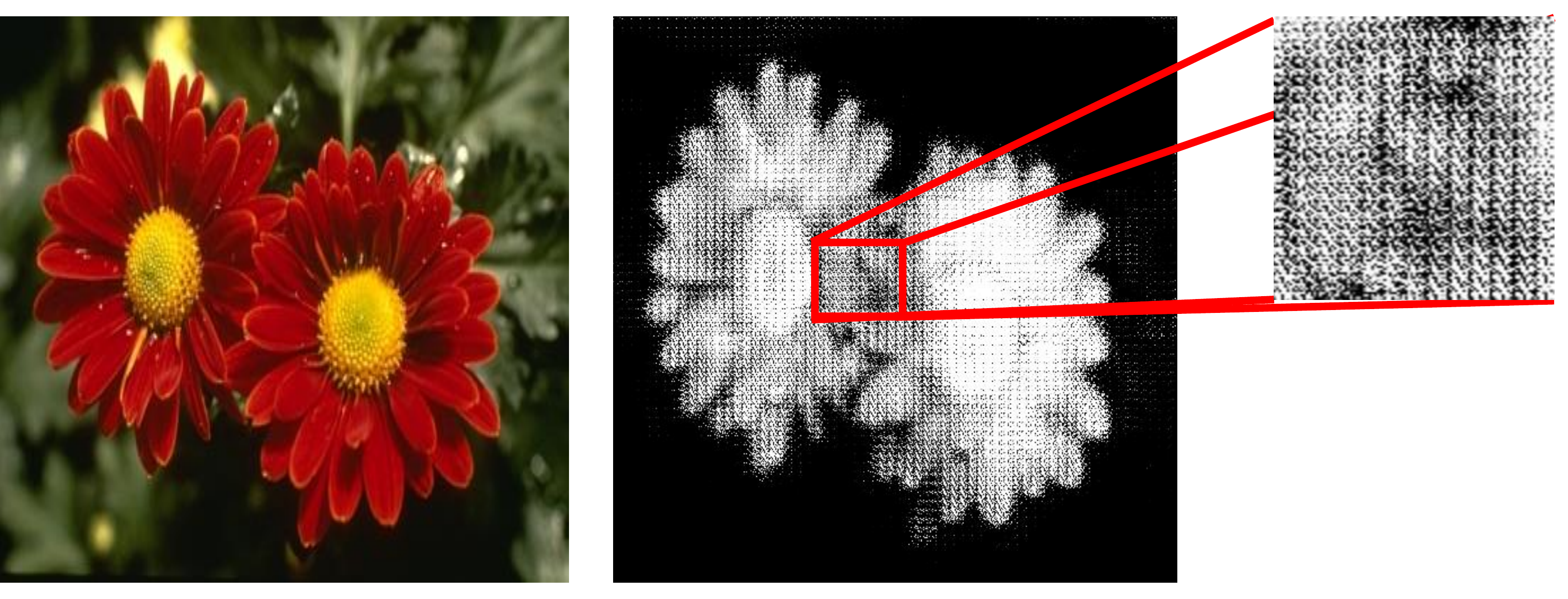}\ \\
{\small(c) Saliency Detection~\cite{wang2016saliency}}\ \\
\includegraphics[width=0.85\linewidth,height=2cm]{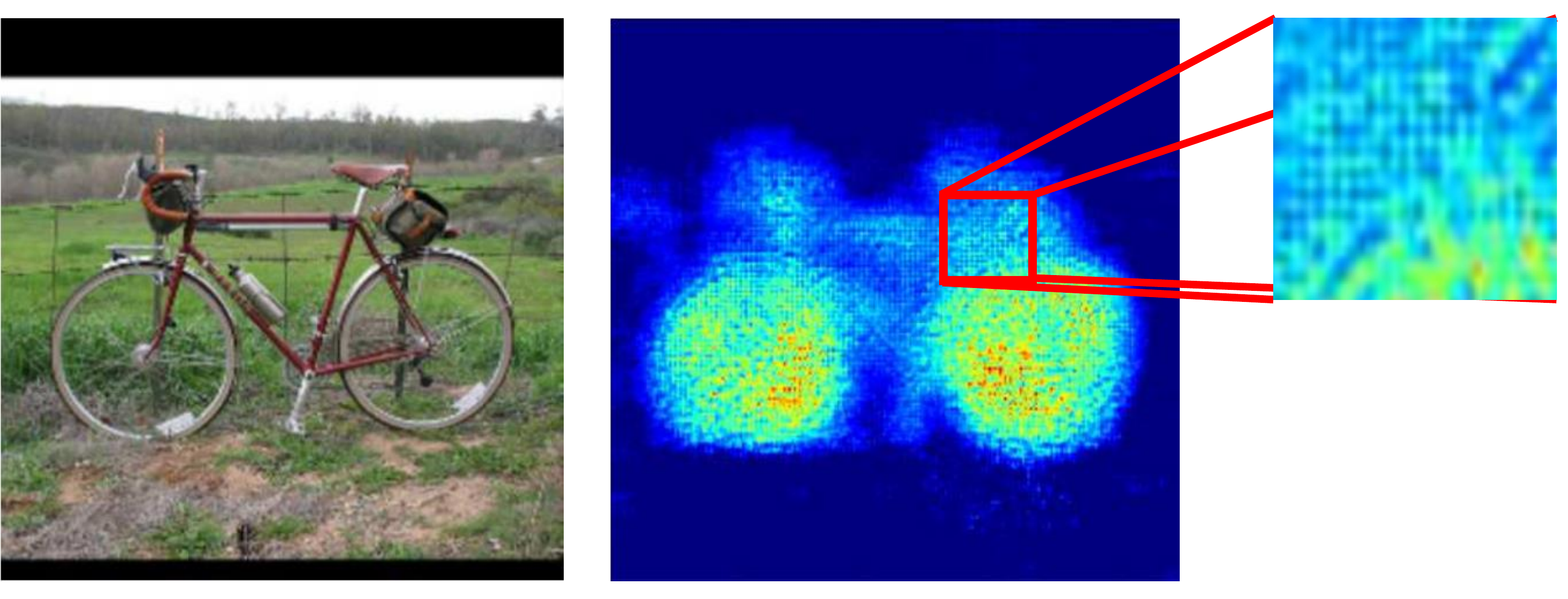}\ \\
{\small(d) Semantic Segmentation~\cite{noh2015learning}}\\
\end{tabular}
\caption{Examples of the checkerboard artifacts in pixel-wise vision tasks using deep CNNs.
High resolution to see better.
\label{fig:checkboard}}
\vspace{-5mm}
\end{figure}
Saliency detection targets to identify the most important and conspicuous objects or regions in an image.
As a pre-processing procedure in computer vision, saliency detection has greatly benefited many practical applications such as object retargeting~\cite{ding2011importance,sun2011scale}, scene classification~\cite{siagian2007rapid}, semantic segmentation~\cite{rother2004grabcut} and visual tracking~\cite{mahadevan2009saliency,hong2015online}.
Although significant progress has been made~\cite{hou2007saliency,alexe2010object,marchesotti2009framework,klein2011center,qin2015saliency,wang2016kernelized}, saliency detection remains very challenging due to complex factors in real world scenarios.
In this work we focus on the task of improving robustness of saliency detection models, which has been ignored in the literature.

Previous saliency detection methods utilize several hand-crafted visual features and heuristic priors.
Recently, deep learning based methods become more and more popular, and have set the benchmark on many datasets~\cite{li2014secrets,yang2013saliency,borji2015salient}.
Their superior performance is partly attributed to the strong representation power in modeling object appearances and varied scenarios.
However, existing methods fail to provide a probabilistic interpretability of the ``black-box'' learning in deep neural networks, and mainly enjoy the models' exceptional performance.
A reasonable probabilistic interpretation can provide relational confidences alongside predictions and make the prediction system into a more robust one~\cite{gal2015dropout}.
In addition, since the uncertainty is a natural part of any predictive system, modeling the uncertainty is of crucial importance.
For instance, the object boundary strongly affects the prediction accuracy of a saliency model, it is desirable that the model can provide meaningful uncertainties on where the boundary of distinct objects is.
As far as we know, there is no work to model and analyze the uncertainty of saliency detection methods based on deep learning.

Another important issue is the checkerboard artifact in pixel-wise vision tasks,
which target to generate images or feature maps from low to high resolution.
Several typical examples are shown in Fig.~\ref{fig:checkboard} (ref.~\cite{odena2016deconvolution} for more details).
The odd artifacts sometimes are very fatal for deep CNNs based approaches.
For example, when the artifacts appear in the output of a fully convolutional network (FCN), the network training may fail and the prediction can be completely wrong~\cite{Shi_2016_CVPR}.
We find that the actual cause of these artifacts is the upsampling mechanism, which generally utilizes the deconvolution operation.
Thus, it is of great interest to explore new upsampling methods to better reduce the artifacts for pixel-wise vision tasks. Meanwhile, the artifacts are also closely related to the uncertainty learning of deep CNNs.

All of the issues discussed above motivate us to learn uncertain features (probabilistic learning) through deep networks to achieve accurate saliency detection. Our model has several unique features, as outlined below.
\begin{itemize}
\vspace{-3mm}
 \item
Different from existing saliency detection methods, our model is extremely simplified.
It consists of an encoder FCN, a corresponding decoder FCN followed by a pixel-wise classification layer.
The encoder FCN hierarchically learns visual features from raw images while the decoder FCN progressively upsamples the encoded feature maps to the input size for the pixel-wise classification.
\vspace{-3mm}
 \item
Our model can learn deep uncertain convolutional features (UCF) for more accurate saliency detection.
The key ingredient is inspired by dropout~\cite{hinton2012improving}.
%
We propose a reformulated dropout (R-dropout), leading to an adaptive ensemble of the internal feature units in specific convolutional layers.
Uncertain features are achieved with no additional parameterization.
 \vspace{-3mm}
 \item
We propose a new upsampling method to reduce the checkerboard artifacts of deconvolution operations.
The new upsampling method has two obvious advantages. On the one hand it separates out upsampling (to generate higher resolution feature maps) from convolution (to extract convolutional features), on the other hand it is compatible with the regular deconvolution.
 \vspace{-3mm}
\item
The uncertain feature extraction and saliency detection are unified in an encoder-decoder network architecture.
The parameters of the proposed model (i.e., weights and biases in all the layers) are jointly trained by end to end gradient learning.
\vspace{-3mm}
\item
Our methods show good generalization on saliency detection and other pixel-wise vision tasks.
Without any post-processing steps, our model yields comparable even better performance on public saliency detection, semantic segmentation and eye fixation datasets.
%
\end{itemize}
\section{Related Work}
Recently, deep learning has delivered superior performance in saliency detection.
For instance, Wang \etal~\cite{wang2015deep} propose two deep neural networks to integrate local estimation and global search for saliency detection.
Li \etal~\cite{li2015visual} train fully connected layers of mutiple CNNs to predict the saliency degree of each superpixel.
To deal with the problem that salient objects may appear in a low-contrast background, Zhao \etal~\cite{zhao2015saliency} take global and local context into account and  model the saliency prediction in a multi-context deep CNN framework.
These methods have excellent performances, however, all of them include fully connected layers, which are very computationally expensive.
What's more, fully connected layers drop spatial information of input images.
To address these issues, Li \etal~\cite{Li2016DeepSaliency} propose a FCN trained under the multi-task learning framework for saliency detection.
Wang \etal~\cite{wang2016saliency} design a recurrent FCN to leverage saliency priors and refine the coarse predictions.
%

Although motivated by the similar spirit, our method significantly differs from~\cite{Li2016DeepSaliency,wang2016saliency} in three aspects.
First, the network architecture is very different. The FCN we used is in the encoder-decoder style, which is in the view of main information reconstruction. In~\cite{Li2016DeepSaliency,wang2016saliency}, the FCN originates from the FCN-8s~\cite{long2015fully} designed with both long and short skip connections for the segmentation task.
Second, instead of simply using FCNs as predictors in~\cite{Li2016DeepSaliency,wang2016saliency}, our model can learn uncertain convolutional features by using multiple reformulated dropouts, which improve the robustness and accuracy of saliency detection.
Third, our model is equipped with a new upsampling method, that naturally handles the checkerboard artifacts of deconvolution operations.
The checkerboard artifacts can be reduced through training the entire neural network.
In contrast, the artifacts is handled by hand-crafted methods in~\cite{Li2016DeepSaliency,wang2016saliency}.
Specifically,~\cite{Li2016DeepSaliency} uses superpixel segmentation to smooth the prediction. 
In~\cite{wang2016saliency}, an edge-aware erosion procedure is used.

Our work is also related to the model uncertainty in deep learning.
%
%
Gal \etal~\cite{gal2015dropout} mathematically prove that a multilayer perceptron models (MLPs) with dropout applied before every weight layer, is equivalent to an approximation to the probabilistic deep Gaussian process.
Though the provided theory is solid, a full verification on deep CNNs is underexplored.
Base on this fact, we make a further step in this direction and show that a reformulated dropout can be used in convolutional layers for learning uncertain feature ensembles.
Another representative work on the model uncertainty is the Bayesian SegNet~\cite{kendall2015bayesian}.
The Bayesian SegNet is able to predict pixel-wise scene segmentation with a measure of the model uncertainty.
They achieve the model uncertainty by Monte Carlo sampling.
Dropout is activated at test time to generate a posterior distribution of pixel class labels.
Different from~\cite{kendall2015bayesian}, our model focuses on learning uncertain convolutional features during training.
%
\begin{figure*}
\begin{center}
\includegraphics[width=0.85\linewidth,height=7.0cm]{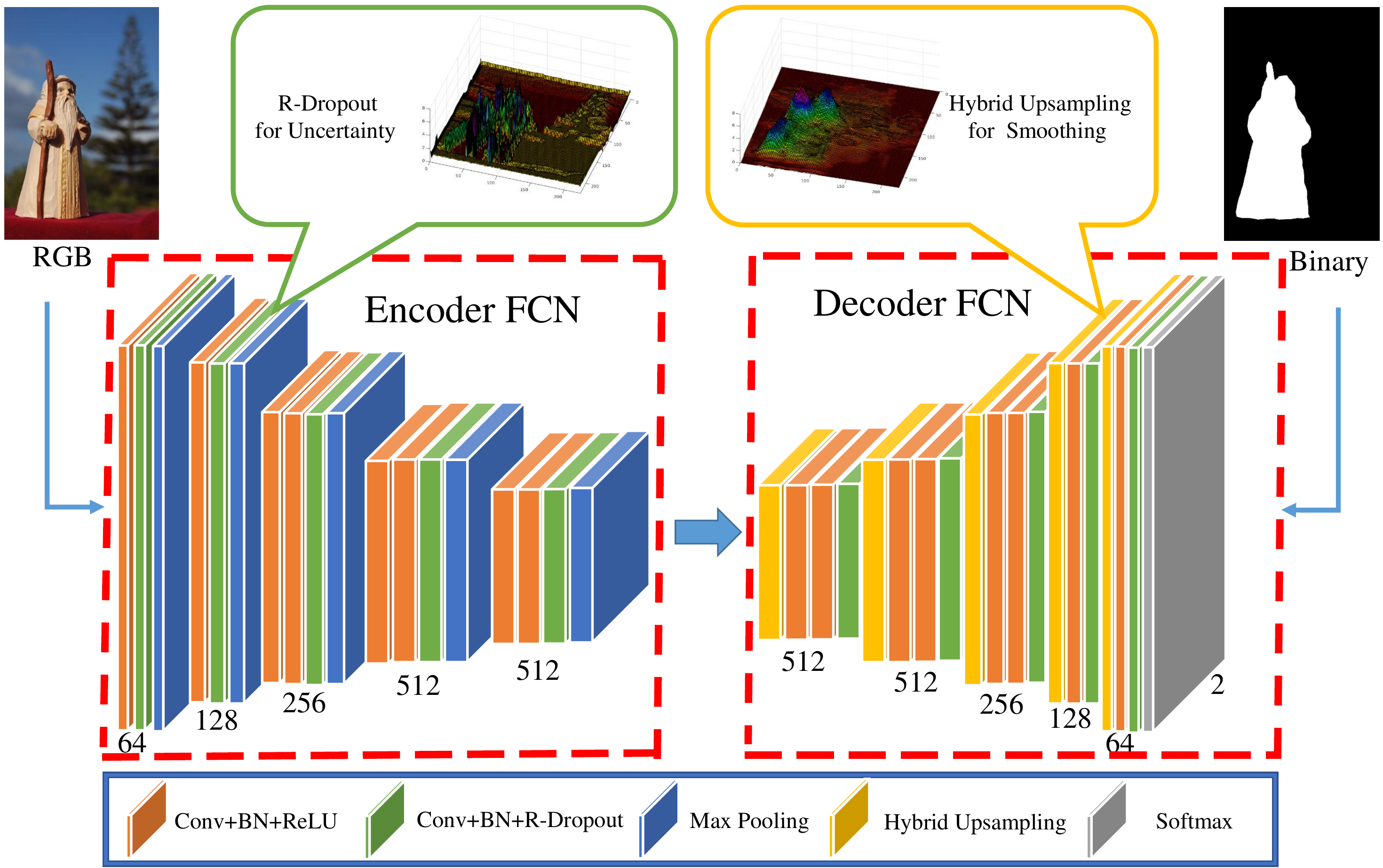}
\end{center}
\vspace{-4mm}
\caption{Overall architecture of the proposed UCF model.}
\label{fig:EDFCN}
\vspace{-5mm}
\end{figure*}
\section{The Proposed Model}
\subsection{Network Architecture Overview}
Our architecture is partly inspired by the stacked denoising auto-encoder~\cite{vincent2010stacked}.
%
We generalize the auto-encoder to a deep fully convolutional encoder-decoder architecture.
The resulting network forms a novel hybrid FCN which consists of an encoder FCN for high-level feature extraction, a corresponding decoder FCN for low-level information reconstruction and a pixel-wise classifier for saliency prediction.
The overall architecture is illustrated in Fig.~\ref{fig:EDFCN}.
More specifically, the encoder FCN consists of multiple convolutional layers with batch normalizations (BN)~\cite{ioffe2015batch} and rectified linear units (ReLU), followed by non-overlapping max pooling.
The corresponding decoder FCN additionally introduces upsampling operations to build feature maps up from low to high resolution. %
We use the \emph{softmax} classifier for the pixel-wise saliency prediction.
In order to achieve the uncertainty of learned convolutional features, we utilize the reformulated dropout (dubbed R-Dropout) after several convolutional layers.
The detailed network configuration is included in supplementary materials.
We will fully elaborate the R-Dropout, our new upsampling method and the training strategy in the following subsections.
\subsection{R-Dropout for Deep Uncertain Convolutional \\Feature Ensemble}
%
Dropout is typically interpreted as bagging a large number of individual models~\cite{hinton2012improving,srivastava2014dropout}.
Although plenty of experiments show that dropout for fully connected layers improves the generalization ability of deep networks, there is a lack of research about using dropout for other type layers, such as convolutional layers.
In this subsection, we show that using modified dropout after convolutional layers can be interpreted as a kind of probabilistic feature ensembles. In light of this fact, we provide a strategy on learning uncertain convolutional features.
{\flushleft\textbf{R-Dropout in Convolution:}} Assume $\textbf{X}\in \mathcal{R}^{W\times H \times C}$ is a 3D tensor, and $f(\textbf{X})$ is a convolution operation in CNNs, projecting $\textbf{X}$ to the $\mathcal{R}^{W^{'}\times H^{'}}$ space by parameters $\textbf{W}$ and $\textbf{b}$:
\vspace{-5mm}
\begin{align}
  f(\textbf{X})=\textbf{WX}+\textbf{b}.
  \label{equ:equ1}
\end{align}
Let $g(\cdot)$ be a non-linear activation function. When the original dropout~\cite{hinton2012improving} is applied to the outputs of $g(f)$, we can get its disturbed version $\hat{g}(f)$ by
\vspace{-1.5mm}
\begin{align}
  g(f) &= g(\textbf{WX}+\textbf{b}),
    \label{equ:equ2}
\end{align}
\vspace{-7mm}
\begin{align}
  \hat{g}(f) &= \textbf{M}\odot g(f)=\textbf{M}\odot g(\textbf{WX}+\textbf{b}),
  \label{equ:equ3}
\end{align}
where $\odot$ denotes element-wise product and $\textbf{M}$ is a binary mask matrix of size $W^{'}\times H^{'}$ with each element $\textbf{M}_{i,j}$ drawn independently from $\textbf{M}_{i,j}\sim Bernoulli(p)$.
Eq.(\ref{equ:equ3}) denotes the activation with dropout during training, and Eq.(\ref{equ:equ2}) denotes the activation at test time.
In addition, Srivastava \etal~\cite{srivastava2014dropout} suggest to scale the activations $g(f)$ with $p$ at test time to obtain an approximate average of the unit activation.

Many commonly used activation functions such as Tanh, ReLU and LReLU~\cite{he2015delving}, have the property that $g(\textbf{0}) = \textbf{0}$.
Thus, Eq.(\ref{equ:equ3}) can be re-written as the R-Dropout formula,
\vspace{-1.5mm}
\begin{align}
  \hat{g}(f) &= g(\textbf{M}\odot(\textbf{WX}+\textbf{b}))\\
             &= g(\textbf{M}\odot(\textbf{WX})+\textbf{M}\odot\textbf{b})\\
             &= g((\textbf{M}\otimes\textbf{W})\textbf{X}+\textbf{M}\odot\textbf{b})\\
             &= g(\textbf{S}\textbf{X}+\textbf{M}\odot\textbf{b}),
  \label{equ:equ7}
\end{align}
where $\otimes$ denotes the cross-channel element-wise product.
From above equations, we can derive that when $\textbf{S} = \textbf{M}\otimes\textbf{W}$ is still binary, Eq.(\ref{equ:equ7}) implies that a kind of stochastic properties\footnote{Stochastic property means that one can use a specific probability distribution to generate the learnable tensor $\textbf{S}$ during each training iteration. The update of $\textbf{S}$ forms a stochastic process not a certain decision.} is applied at the inputs to the activation function.
Let $\textbf{S}_{i,j,k}\in [0,1]$ and $\sum_{j}\sum_{k}\textbf{S}_{i,j,k}=1$, the above equations will strictly construct an ensemble of internal feature units of $\textbf{X}$.
However, in practice there is certainly no evidence to hold above constraints.
Even though, we note that:
(1) the stochastic mask matrix $\textbf{S}$ mainly depends on the mask generator\footnote{In R-Dropout, the generator can be any probability distribution. The original dropout is a special case of the R-Dropout, when the generator is the Bernoulli distribution.};
(2) when training deep convolutional networks, the R-Dropout after convolutional layers acts as an uncertain ensemble of convolutional features;
(3) this kind of feature ensemble is element-wisely probabilistic, thus it can bring forth robustness in the prediction of dense labeling vision tasks such as saliency detection and semantic segmentation.
\vspace{-5mm}
{\flushleft\textbf{Uncertain Convolutional Feature Extraction:}} Motivated by above insights, we employ the R-Dropout into convolutional layers of our model, thereby generating deep uncertain convolutional feature maps.
Since our model consists of alternating convolutional and pooling layers, there exist two typical cases in our model.
For notational simplicity, we subsequently drop the batch normalization (BN).

\textbf{1) Conv+R-Dropout+Conv:}
If the proposed R-Dropout is followed by a convolutional layer, the forward propagation of input is formulated as
\begin{align}
  \hat{g}(f^{l}) &= g(\textbf{S}^{l}\textbf{X}^{l-1}+\textbf{M}^{l}\odot\textbf{b}^{l}),
    \label{equ:equ8}
\end{align}
\vspace{-7mm}
\begin{align}
f^{l+1} &= \textbf{Conv}(\textbf{W}^{l+1},\hat{g}(f^{l})),
  \label{equ:equ9}
\end{align}
\vspace{-7mm}
\begin{align}
g^{l+1}= g(f^{l+1}),
  \label{equ:equ10}
\end{align}
where $l$ is the layer number and $\textbf{Conv}$ is the convolution operation. As we can see from Eq.(\ref{equ:equ9}), the disturbed activation $\hat{g}(f^{l})$ is convolved with filter $\textbf{W}^{l+1}$ to produce convolved features $f^{l+1}$.
In this way, the network will focus on learning the weight and bias parameters, i.e., $\textbf{W}$ and $\textbf{b}$, and the uncertainty of using the R-Dropout will be dissipated during training deep networks.

\textbf{2) Conv+R-Dropout+Pooling:}
In this case, the forward propagation of input becomes
\begin{align}
  \hat{g}(f^{l}) &= g(\textbf{S}^{l}\textbf{X}^{l-1}+\textbf{M}^{l}\odot\textbf{b}^{l}),
\end{align}
\vspace{-7mm}
\begin{align}
&g^{l+1}_{j}=\textbf{Pooling}(\hat{g}(f^{l})_{1},...,\hat{g}(f^{l})_{n}),i\in R^{l}_j.&
\label{equ:equ12}
\end{align}
Here $\textbf{Pooling}(\cdot)$ denotes the max-pooling function. $R^{l}_j$ is the pooling region $j$ at layer $l$ and $\hat{g}(f^{l})_{i}$ is the activition of each neuron within $R^{l}_j$. $n=|R^{l}_{j}|$ is the number of units in $R^{l}_j$. To formulate the uncertainty, without loss of generality, we suppose the activations $\hat{g}(f^{l})$ in each pooling region $j$ are ordered in non-decreasing order, i.e. $\hat{g}(f^{l})_{1}\leq \hat{g}(f^{l})_{2}\leq...\leq\hat{g}(f^{l})_{n}$. As a result, $\hat{g}(f^{l})_{i}$ will be selected as the pooled activation on conditions that (1) $\hat{g}(f^{l})_{i+1}, \hat{g}(f^{l})_{i+2},...,\hat{g}(f^{l})_{n}$ are dropped out, and (2) $\hat{g}(f^{l})_{i}$ is retained.
This event occurs with probability of $P_{i}$ according to the probability theory,
\begin{align}
  P(g^{l+1}_{j}=\hat{g}(f^{l})_{i}) &= P_{i}=pq^{n-i}, p=1-q.
  \label{equ:equ13}
\end{align}
Therefore, performing R-dropout before the max-pooling operation is exactly sampling from the following multinomial distribution
to select an index $i$, then the pooled activation is simply $\hat{g}(f^{l})_{i}$,
\begin{align}
  g^{l+1}_{j}=\hat{g}(f^{l})_{i}, i\sim Multinomial( P_{0}, P_{1},..., P_{n}),
  \label{equ:equ14}
\end{align}
where $P_{0}(=q^{n})$ is the special event that all the units in a pooling region is dropped out.

The latter strategy exhibits the effectiveness of building the uncertainty by employing the R-Dropout into convolutional layers.
We adopt it to build up our network architecture (see Fig.~\ref{fig:EDFCN}).
We will experimentally demonstrate that the R-Dropout based FCN yields marvelous results on the saliency detection datasets in Section 4.
\subsection{Hybrid Upsampling for Prediction Smoothing}
In this subsection, we first explicate the cause of checkerboard artifacts by the deconvolution arithmetic~\cite{vincent2016arithmetic}.
Then we derive a new upsampling method to reduce the artifacts as much as possible for the network training and inference.

\begin{figure}
\centering
\begin{tabular}{@{}c@{}c@{}c@{}c}
\includegraphics[width=0.25\linewidth,height=2.5cm]{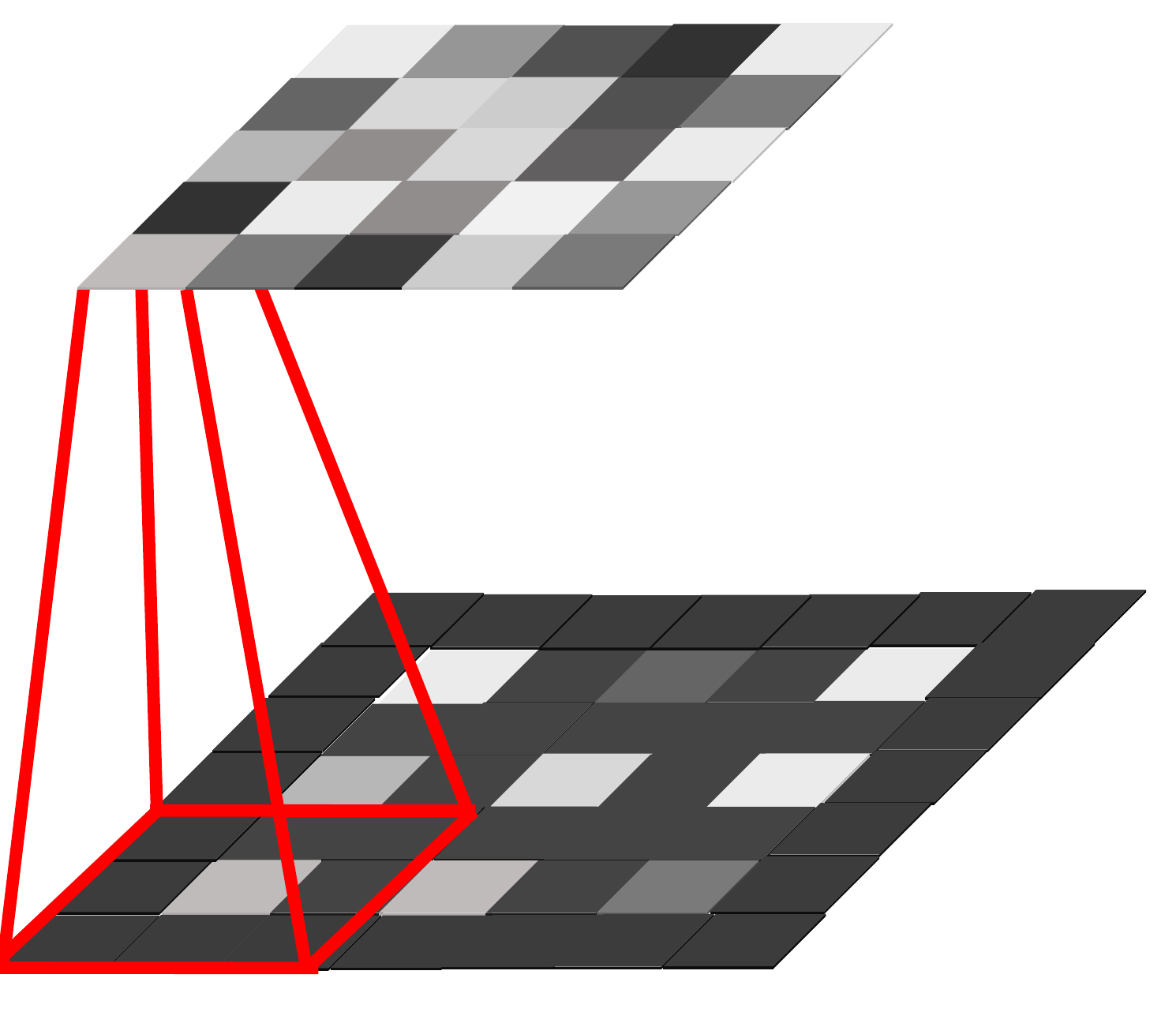}\ &
\includegraphics[width=0.25\linewidth,height=2.5cm]{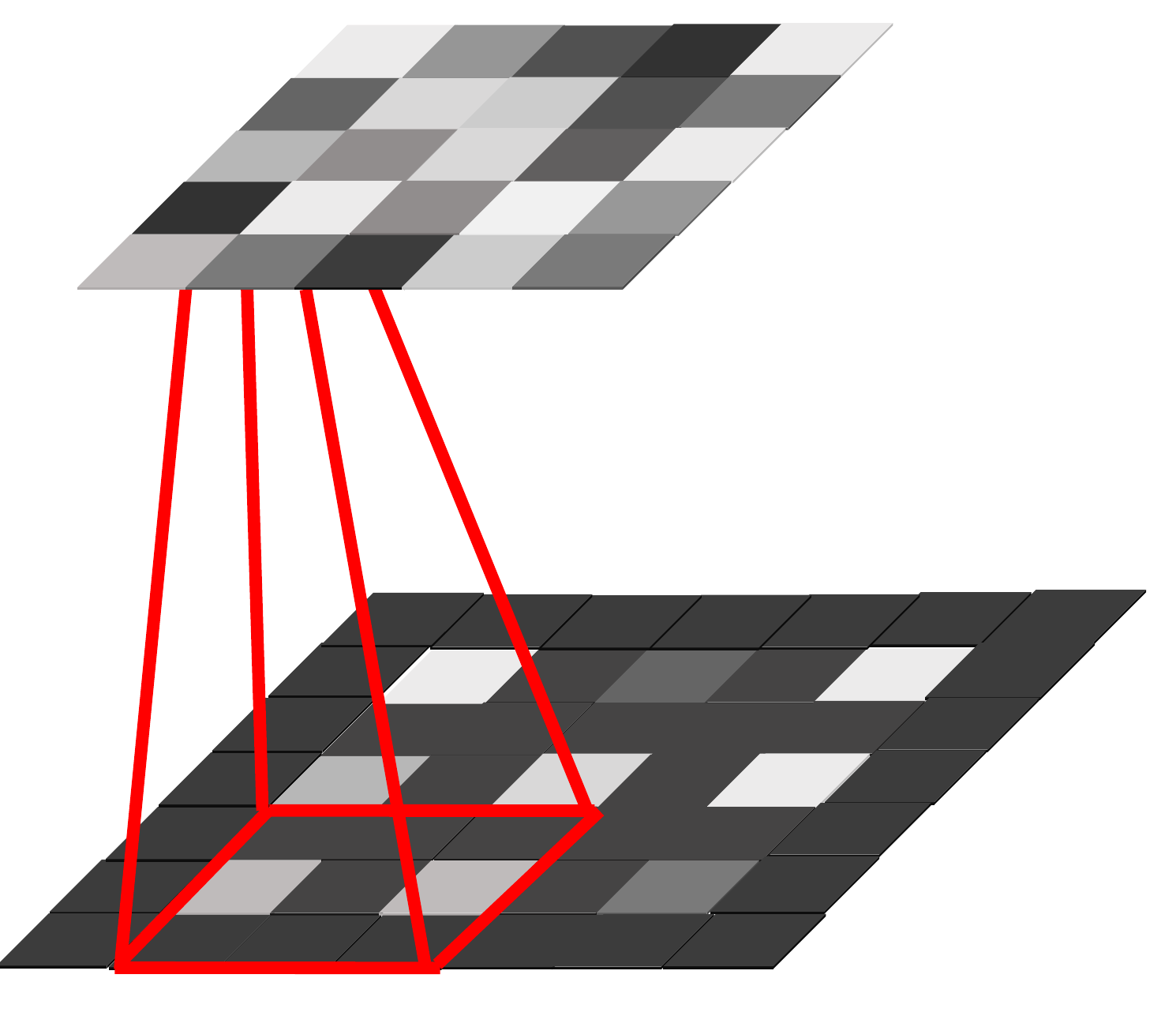}\ &
\includegraphics[width=0.25\linewidth,height=2.5cm]{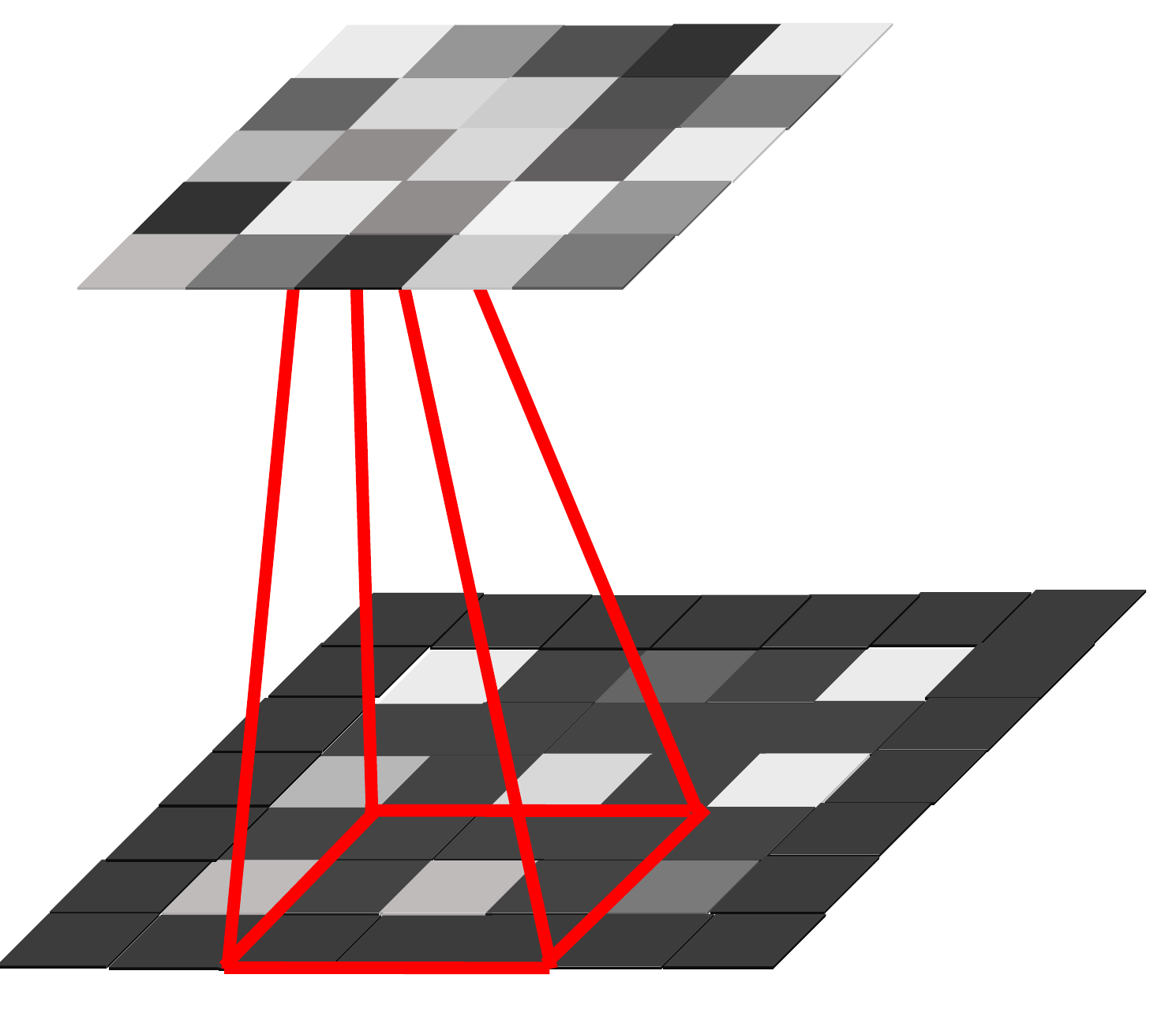}\ &
\includegraphics[width=0.25\linewidth,height=2.5cm]{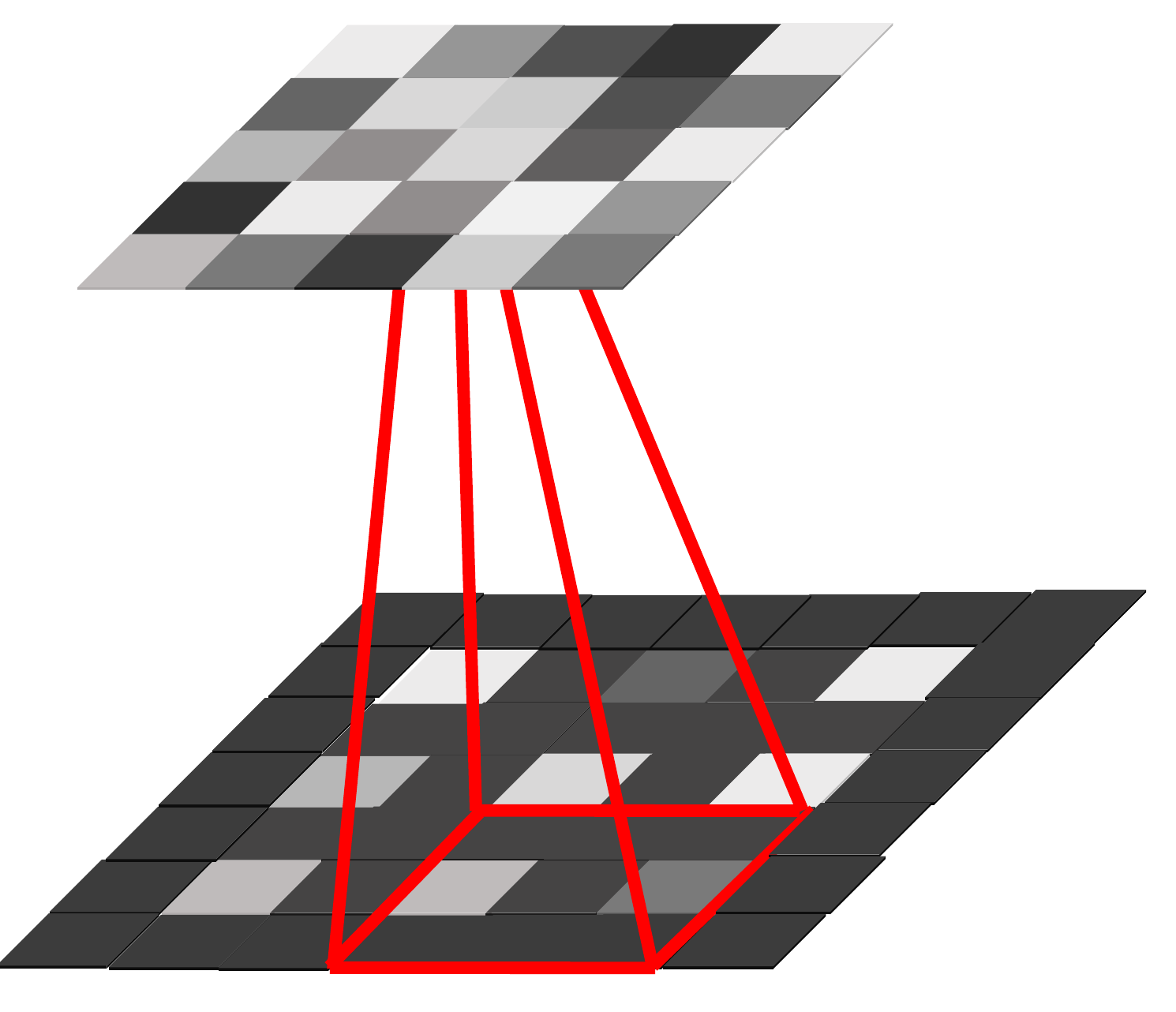}\
\end{tabular}
\\
\caption{The detailed explanation of deconvolution. The deconvolution of a $3\times3$ kernel using $2\times2$ strides over a $5\times5$ input padded with a $1\times1$ border of zeros (i.e., $n = 5$, $k = 3$, $s = 2$ and $p = 1$).
It is equivalent to convolving a $3\times3$ kernel over a $3\times3$ input (with 1 zero inserted between inputs) padded with a $1\times1$ border of zeros using unit strides (i.e., $n^{'} = 3$, $\hat{n}^{'} = 5$, $k = 3$, $s = 1$ and $p = 1$).
}
\label{fig:Deconv}
\vspace{-3mm}
\end{figure}

Without loss of generality, we focus on the square input ($n\times n$), square kernel size ($k\times k$), same stride ($s\times s$) and same zero padding ($p\times p$) (if used) along both axes.
Since we aim to implement upsampling, we set $s\ge2$.
In general, the convolution operation $\bf{C}$ can be described by
\begin{align}
  \bf{O = C(I,F)=I*F},
  \label{equ:equ15}
\end{align}
where $\textbf{I}_{(n+p)\times(n+p)}$ is the input, $\textbf{F}_{k\times k}$ is the filter with stride $s$, $*$ is the discrete convolution and $\textbf{O}$ is the output whose dimension is $\lfloor (n+2p-k)/s\rfloor+1$.
The convolution $\bf{C}$ has an associated deconvolution $\bf{D}$ described by $\hat{n}^{'}$, $k^{'}=k$, $s^{'}=1$ and $p^{'}=k-p-1$, where $\hat{n}^{'}$ is the size of the stretched input obtained by adding $s-1$ zeros between each input unit, and the output size of the deconvolution is $o^{'}=s(n^{'}-1)+k-2p$\footnote {The constraint on the size of the input $n$ can be relaxed by introducing another parameter $t\in{0,...,s-1}$ that allows to distinguish between the $s$ different cases that all lead to the same $n^{'}$.}.
This indicates that \textbf{the regular deconvolution operator is equivalent to performing convolution on a new input with inserted zeros} ($\hat{n}^{'}\times\hat{n}^{'}$).
A toy example is shown in Fig.~\ref{fig:Deconv}.
In addition, when the filter size $k$ can not be divided by the stride $s$, the deconvolution will cause the overlapping issue.
If the stretched input is high-frequency or near periodic, i.e., the value is extremely undulating when zeros are inserted, the output results of deconvolution operations naturally have numerical artifacts like a checkerboard.
\begin{figure}
\begin{center}
  \includegraphics[width=0.9\linewidth,height=5.0cm]{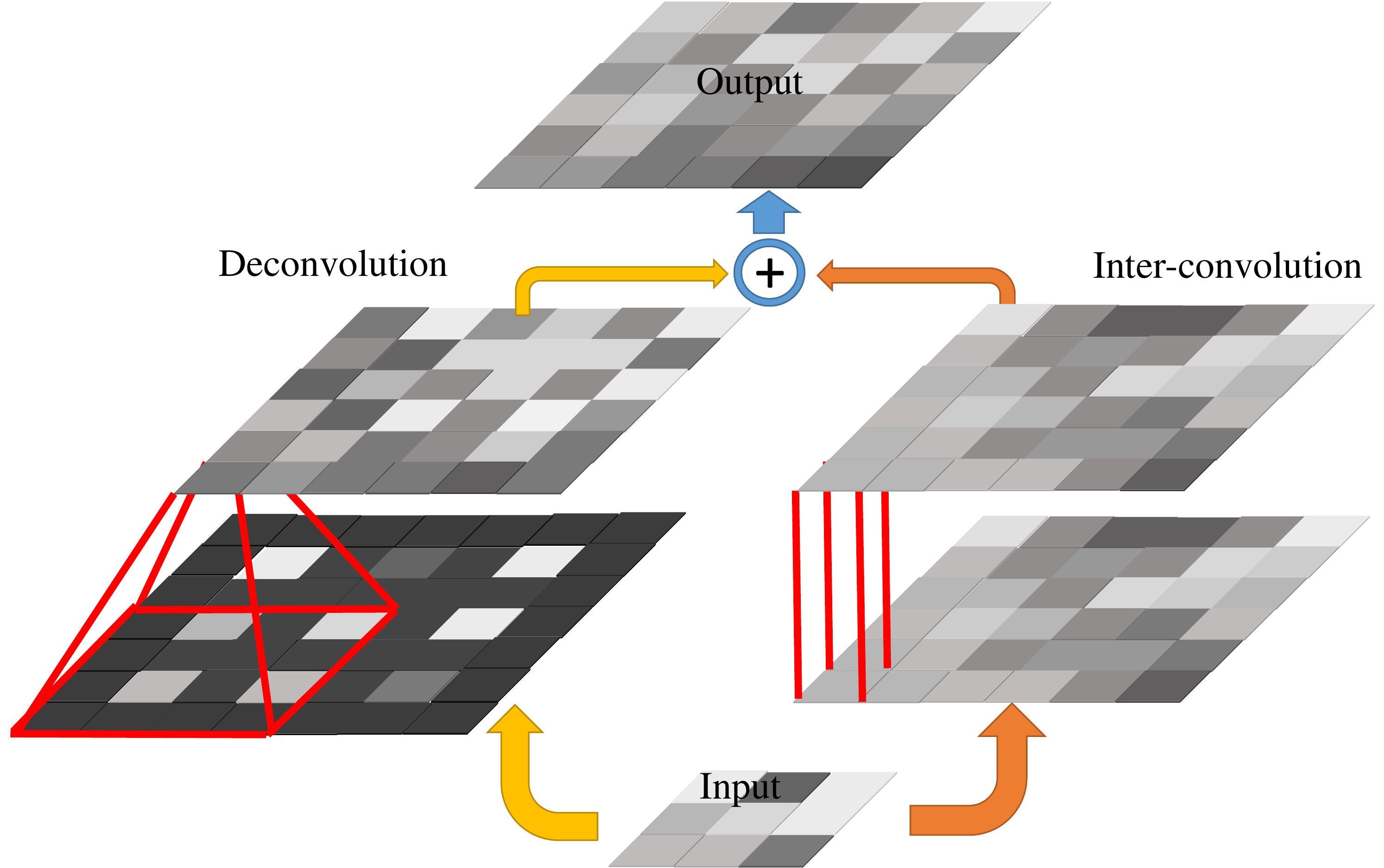}
\end{center}
\vspace{-3mm}
\caption{The hybrid upsampling. Two strategies are jointly used: 1) deconvolution with restricted filter sizes (left branch); 2) linear interpolation with 1x1 convolution (right branch). The final output is the summed result of two strategies.}
\label{fig:upsampling}
\vspace{-5mm}
\end{figure}

Base on the above observations, we propose two strategies to avoid the artifacts produced by the regular deconvolution.
The first one is restricting the filter size. We can simply ensure the filter size is a multiple of the stride size, avoiding the overlapping issue, i.e.,
\begin{align}
  k=\lambda s, \lambda\in \mathcal{N_{+}}.
  \label{equ:equ16}
\end{align}
Then the deconvolution will dispose the zero-inserted input with the equivalent convolution, deriving a smooth output.
However, because this method only focuses on changing the receptive fields of the output, and can not change the frequency distribution of the zero-inserted input, the artifacts can still leak through in several extreme cases.
We propose another alternative strategy which separates out upsampling from equivalent convolution.
We first resize the original input into the desired size by interpolations, and then perform some equivalent convolutions.
Although this strategy may destroy the learned features in deep CNNs, we find that high resolution maps built by iteratively stacking this kind of upsampling can reduce artifacts amazingly.
In order to take the strength of both strategies, we introduce the hybrid upsampling method by summing up the outputs of the two strategies.
Fig.~\ref{fig:upsampling} illustrates the proposed upsampling method.
In our proposed model, we use bilinear (or nearest-neighbor) operations for the interpolation.
These interpolation methods are linear operations, and can be embedded into the deep CNNs as efficient matrix multiplications.
\subsection{Training the Entire Network}
Since there is a lack of enough saliency detection data for training our model from scratch, we utilize the front-end of the VGG-16 model~\cite{simonyan2014very} as our encoder FCN (13 convolutional layers and 5 pooling layers pre-trained on ILSVRC 2014 for the image classification task).
Our decoder FCN is a mirrored version of the encoder FCN, and has multiple series of upsampling, convolution
and rectification layers.
Batch normalization (BN) is added to the output of every convolutional layer.
%
We add the R-dropout with an equal sampling rate $p = 0.5$ after specific convolutional layers, as shown in Fig.~\ref{fig:EDFCN}.
For saliency detection, we randomly initialize the weights of the decoder FCN and fine-tune the entire network on the MSRA10K dataset~\cite{ChengPAMI}, which is widely used in salient object detection community (More details will be described in Section 4).
We convert the ground-truth saliency map of each image in that dataset to be a 0-1 binary map.
This kind of transform perfectly matches the channel output of the FCN when we use the \emph{softmax} cross-entropy loss function given by the following equation~(\ref{equ:equ17}) for separating saliency foreground from general background.
\begin{align}
  \mathcal{L} & = -\sum\limits_{m}l_{m}\log(q_m)+(1-l_{m})\log(1-q_m),
  \label{equ:equ17}
\end{align}
where $l_{m}$ $(=0,1)$ is the label of a pixel $m$ in the image and $q_m$ is the probability that the pixel is the saliency foreground.
The value of $q_m$ is obtained from the output of the network.
Before putting the training images into our proposed model, each image is subtracted with the ImageNet mean~\cite{imagenet_cvpr09} and rescaled into the same size (448 $\times$ 448).
For the correspondence, we also rescale the 0-1 binary maps to the same size.
The model is trained end to end using the mini-batch stochastic gradient descent (SGD) with a momentum, learning rate decay schedule.
The detailed settings of parameters are included in the supplementary material.
\subsection{Saliency Inference}
Because our model is a fully convolutional network, it can take images with arbitrary size as inputs when testing.
After the feed-forward process, the output of the network is composed of a foreground excitation map ($\textbf{M}^{fe}$) and a background excitation map ($\textbf{M}^{be}$).
We use the difference between $\textbf{M}^{fe}$ and $\textbf{M}^{be}$, and clip the negative values to obtain the resulting saliency map, i.e.,
\begin{align}
  \textbf{Sal} & = max(\textbf{M}^{fe}-\textbf{M}^{be},0).
  \label{equ:equ18}
\end{align}
This subtraction strategy not only increases the pixel-level discrimination but also captures context contrast information.
Optionally, we can take the ensemble of multi-scale predicted maps to further improve performance.
\section{Experiments}
In this section, we start by describing the experimental setup for saliency detection.
Then, we thoroughly evaluate and analyze our proposed model on public saliency detection datasets.
Finally, we provide additional experiments to verify the generalization of our methods on other pixel-wise vision tasks, i.e., semantic segmentation and eye fixation.
\subsection{Experimental Setup}
\textbf{Saliency Datasets:} For training the proposed network, we simply augment the MSRA10K dataset~\cite{ChengPAMI} by the mirror reflection and rotation techniques ($0^{\circ}, 90^{\circ}, 180^{\circ}, 270^{\circ}$), producing 80,000 training images totally. %

For the detection performance evaluation, we adopt six widely used saliency detection datasets as follows,

\textbf{DUT-OMRON}~\cite{yang2013saliency}. This dataset consists of 5,168 high quality images. Images in this dataset have one or more salient objects and relatively complex background. Thus, this dataset is difficult and challenging in saliency detection.

\textbf{ECSSD}~\cite{yan2013hierarchical}. This dataset contains 1,000 natural images, including many semantically meaningful and complex structures in the ground truth segmentations.

\textbf{HKU-IS}~\cite{zhao2015saliency}. This dataset contains 4,447 images with high quality pixel-wise annotations.
Images in this dataset are well chosen to include multiple disconnected objects or objects touching the image boundary.

\textbf{PASCAL-S}~\cite{li2014secrets}. This dataset is carefully selected from the PASCAL VOC dataset~\cite{Everingham2010ThePV} and contains 850 images.

\textbf{SED}~\cite{borj2015salient}. This dataset contains two different subsets: \textbf{SED1} and \textbf{SED2}.
The \textbf{SED1} has 100 images each containing only one salient object, while the \textbf{SED2} has 100 images each containing two salient objects.

\textbf{SOD}~\cite{yan2013hierarchical}. This dataset has 300 images, and it was originally designed for image segmentation.
Pixel-wise annotation of salient objects was generated by~\cite{jiang2013salient}.
%

\textbf{Implementation Details:}
We implement our approach based on the MATLAB R2014b platform with the modified Caffe toolbox~\cite{kendall2015bayesian}.
We run our approach in a quad-core PC machine with an i7-4790 CPU (with 16G memory) and one NVIDIA Titan X GPU (with 12G memory).
The training process of our model takes almost 23 hours and converges after 200k iterations of the min-batch SGD.
The proposed saliency detection algorithm runs at about \textbf{7 fps} with $448\times448$ resolution (\textbf{23 fps} with $224\times224$ resolution).
The source code can be found at \textcolor[rgb]{1,0,0}{http://ice.dlut.edu.cn/lu/}.

\textbf{Saliency Evaluation Metrics:}
We adopt three widely used metrics to measure the performance of all algorithms, i.e., the Precision-Recall (PR) curves, F-measure and Mean Absolute Error (MAE)~\cite{borji2015salient}.
The precision and recall are computed by thresholding the predicted saliency map, and comparing the binary map with the ground truth.
The PR curve of a dataset indicates the mean precision and recall of saliency maps at different thresholds.
The F-measure is a balanced mean of average precision and average recall, and can be calculated by
\begin{align}
  F_{\beta} =\frac{(1+\beta^2)\times Precision\times Recall}{\beta^2\times Precision \times Recall}.
    \label{equ:equ19}
\end{align}
Following existing works~\cite{yan2013hierarchical}~\cite{wang2015deep}~\cite{borji2015salient}~\cite{yang2013saliency}, we set $\beta^2$ to be 0.3 to weigh precision more than recall.
We report the performance when each saliency map is adaptively binarized with an image-dependent threshold.
The threshold is determined to be twice the mean saliency of the image:
\begin{align}
T = \frac{2}{W\times H}\sum_{x=1}^{W}\sum_{y=1}^{H}S(x,y),
  \label{equ:equ20}
\end{align}
where $W$ and $H$ are width and height of an image, $S(x,y)$ is the saliency value of the pixel at
$(x,y)$.
%
%
%

We also calculate the mean absolute error (MAE) for fair comparisons as suggested by~\cite{borji2015salient}.
The MAE evaluates the saliency detection accuracy by
\vspace{-1mm}
\begin{align}
MAE = \frac{1}{W\times H}\sum_{x=1}^{W}\sum_{y=1}^{H}|S(x,y)-G(x,y)|,
  \label{equ:equ21}
\end{align}
\vspace{-0.5mm}
where $G$ is the binary ground truth mask.
\begin{figure*}
\begin{center}
\begin{tabular}{@{}c@{}c@{}c@{}c@{}c@{}c@{}c}
\includegraphics[width=0.285\linewidth,height=3.75cm]{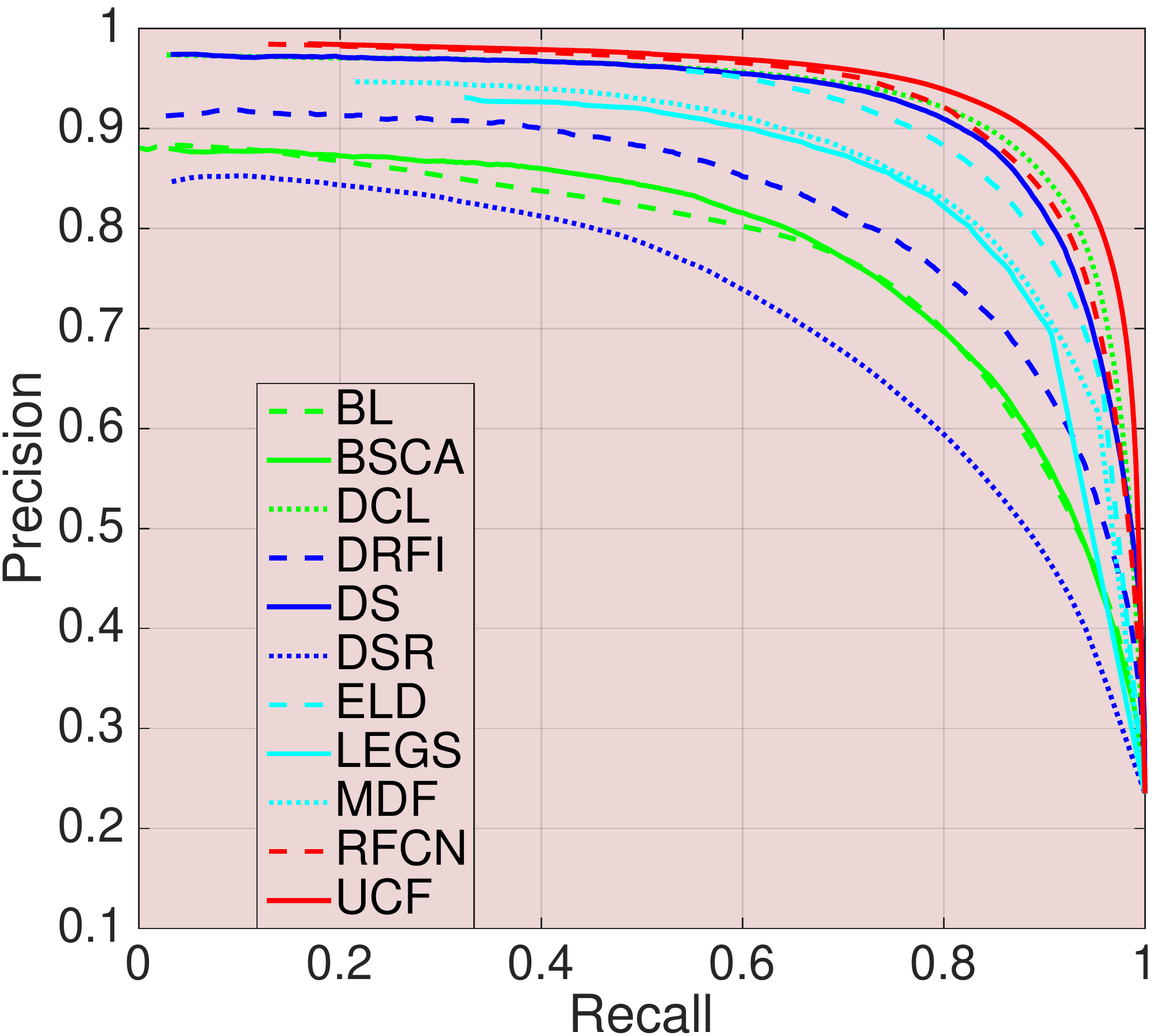} \ \quad\quad   &
\includegraphics[width=0.285\linewidth,height=3.75cm]{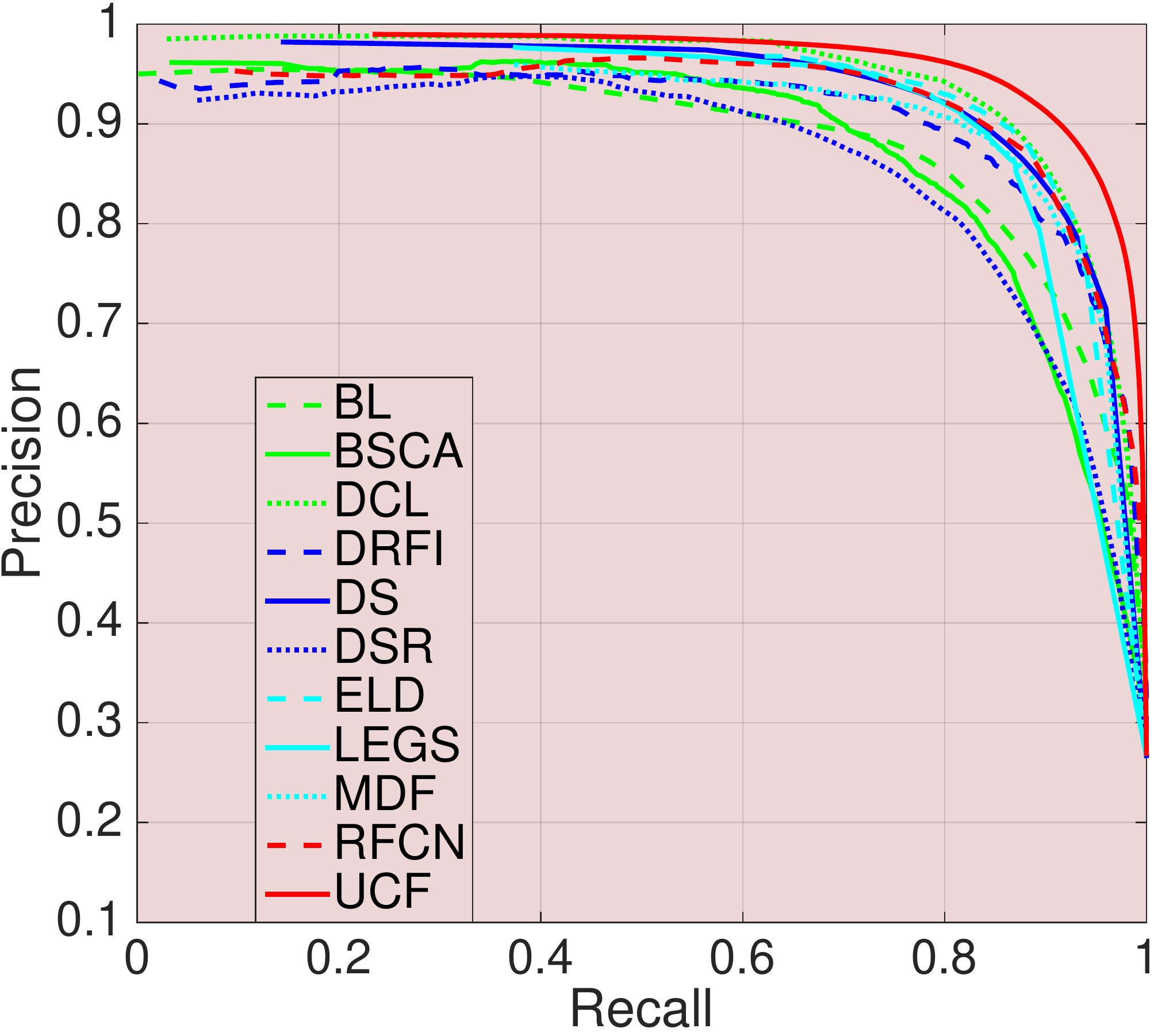} \ \quad\quad   &
\includegraphics[width=0.285\linewidth,height=3.75cm]{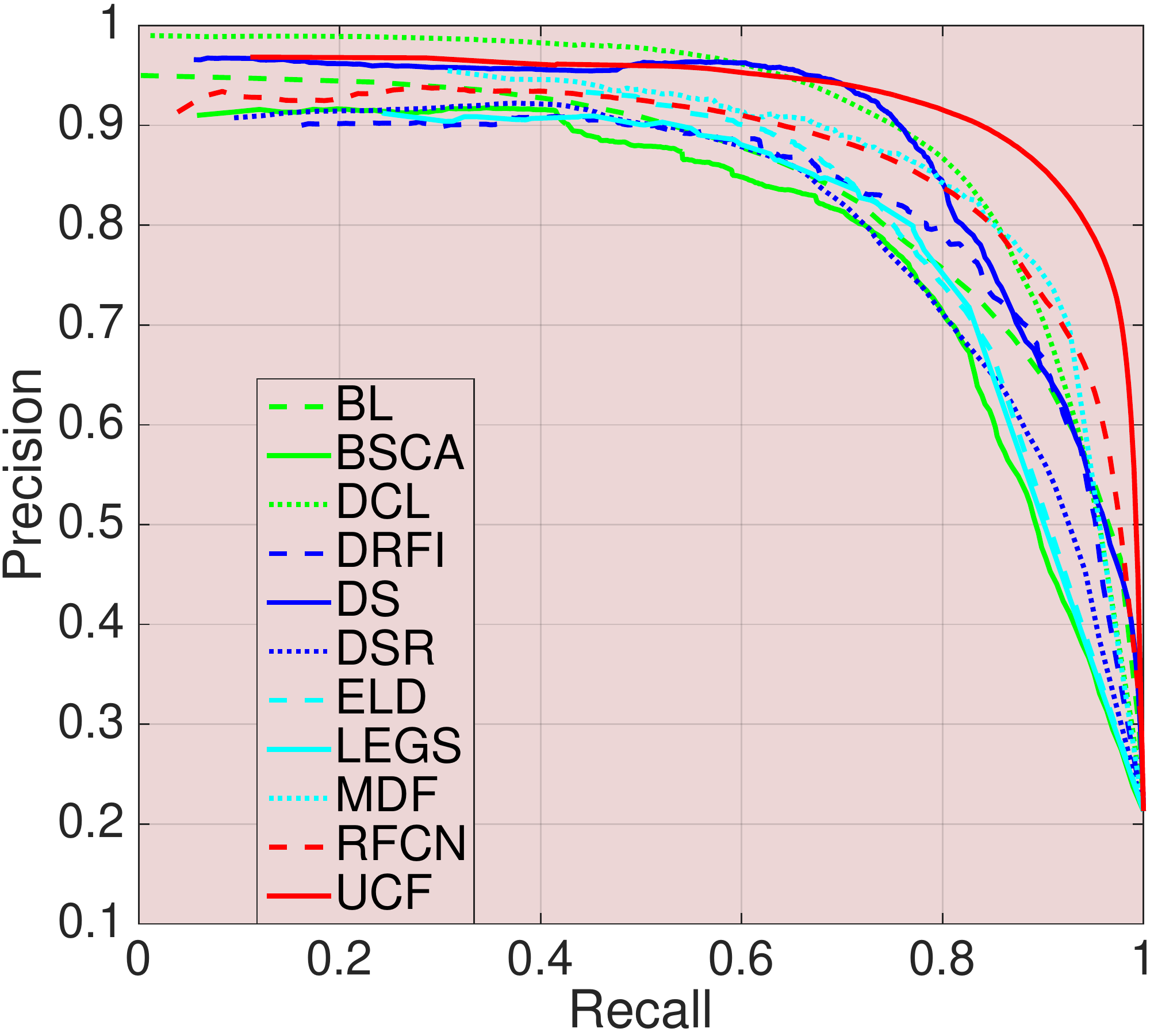} \ \\
 {\small(a) ECSSD} & {\small(b) SED1} & {\small(c) SED2} \ \\
\end{tabular}
\caption{Performance of the proposed algorithm compared with other state-of-the-art methods.
\label{fig:PR-curve}}
\end{center}
\end{figure*}
\setlength{\tabcolsep}{2.75pt}
\begin{table*}
\vspace{-4mm}
\begin{center}
\doublerulesep=0.6pt
\begin{tabular}{|c|c|c|c|c|c|c|c|c|c|c|c|c|c|c||c|c|c|c|c|c|c|c|c|c|c|c|c|c|c|c|c|c|c|c|c|c|c|c|c|||c|c|c|c|c|c|c|c|||}
\hline
\multicolumn{4}{|c|}{}
&\multicolumn{4}{|c|}{DUT-OMRON}
&\multicolumn{4}{|c|}{ECSSD}
&\multicolumn{4}{|c|}{HKU-IS}
&\multicolumn{4}{|c|}{PASCAL-S}
&\multicolumn{4}{|c|}{SED1}
&\multicolumn{4}{|c|}{SED2}
\\
\hline
\multicolumn{4}{|c|}{Methods}
&\multicolumn{2}{|c|}{$F_\beta$}&\multicolumn{2}{|c|}{$MAE$}
&\multicolumn{2}{|c|}{$F_\beta$}&\multicolumn{2}{|c|}{$MAE$}
&\multicolumn{2}{|c|}{$F_\beta$}&\multicolumn{2}{|c|}{$MAE$}
&\multicolumn{2}{|c|}{$F_\beta$}&\multicolumn{2}{|c|}{$MAE$}
&\multicolumn{2}{|c|}{$F_\beta$}&\multicolumn{2}{|c|}{$MAE$}
&\multicolumn{2}{|c|}{$F_\beta$}&\multicolumn{2}{|c|}{$MAE$}
\\
\hline
\multicolumn{4}{|c|}{UCF}
&\multicolumn{2}{|c|}{\textcolor[rgb]{0,0,1}{0.6283}}&\multicolumn{2}{|c|}{0.1203}
&\multicolumn{2}{|c|}{\textcolor[rgb]{1,0,0}{0.8517}}&\multicolumn{2}{|c|}{\textcolor[rgb]{1,0,0}{0.0689}}
&\multicolumn{2}{|c|}{\textcolor[rgb]{0,0,1}{0.8232}}&\multicolumn{2}{|c|}{\textcolor[rgb]{1,0,0}{0.0620}}
&\multicolumn{2}{|c|}{\textcolor[rgb]{0,1,0}{0.7413}}&\multicolumn{2}{|c|}{\textcolor[rgb]{1,0,0}{0.1160}}
&\multicolumn{2}{|c|}{\textcolor[rgb]{1,0,0}{0.8647}}&\multicolumn{2}{|c|}{\textcolor[rgb]{1,0,0}{0.0631}}
&\multicolumn{2}{|c|}{\textcolor[rgb]{1,0,0}{0.8102}}&\multicolumn{2}{|c|}{\textcolor[rgb]{1,0,0}{0.0680}}
\\
\multicolumn{4}{|c|}{V-E}
&\multicolumn{2}{|c|}{0.6135}&\multicolumn{2}{|c|}{0.1224}
&\multicolumn{2}{|c|}{0.7857}&\multicolumn{2}{|c|}{0.0795}
&\multicolumn{2}{|c|}{0.7716}&\multicolumn{2}{|c|}{0.0785}
&\multicolumn{2}{|c|}{0.6303}&\multicolumn{2}{|c|}{0.1284}
&\multicolumn{2}{|c|}{0.8128}&\multicolumn{2}{|c|}{0.0732}
&\multicolumn{2}{|c|}{0.7576}&\multicolumn{2}{|c|}{0.0851}
\\
\multicolumn{4}{|c|}{V-D}
&\multicolumn{2}{|c|}{0.5072}&\multicolumn{2}{|c|}{0.1345}
&\multicolumn{2}{|c|}{0.6942}&\multicolumn{2}{|c|}{0.1195}
&\multicolumn{2}{|c|}{0.6851}&\multicolumn{2}{|c|}{0.0967}
&\multicolumn{2}{|c|}{0.5695}&\multicolumn{2}{|c|}{0.1624}
&\multicolumn{2}{|c|}{0.7754}&\multicolumn{2}{|c|}{0.0844}
&\multicolumn{2}{|c|}{0.6930}&\multicolumn{2}{|c|}{0.0954}
\\
\multicolumn{4}{|c|}{V-C}
&\multicolumn{2}{|c|}{0.6165}&\multicolumn{2}{|c|}{0.1210}
&\multicolumn{2}{|c|}{\textcolor[rgb]{0,1,0}{0.8426}}&\multicolumn{2}{|c|}{\textcolor[rgb]{0,1,0}{0.0711}}
&\multicolumn{2}{|c|}{0.8156}&\multicolumn{2}{|c|}{\textcolor[rgb]{0,0,1}{0.0670}}
&\multicolumn{2}{|c|}{\textcolor[rgb]{0,0,1}{0.7201}}&\multicolumn{2}{|c|}{\textcolor[rgb]{0,1,0}{0.1203}}
&\multicolumn{2}{|c|}{\textcolor[rgb]{0,1,0}{0.8665}}&\multicolumn{2}{|c|}{\textcolor[rgb]{0,1,0}{0.0653}}
&\multicolumn{2}{|c|}{\textcolor[rgb]{0,1,0}{0.8014}}&\multicolumn{2}{|c|}{\textcolor[rgb]{0,0,1}{0.0795}}
\\
\multicolumn{4}{|c|}{V-B}
&\multicolumn{2}{|c|}{0.6168}&\multicolumn{2}{|c|}{0.1305}
&\multicolumn{2}{|c|}{\textcolor[rgb]{0,0,1}{0.8356}}&\multicolumn{2}{|c|}{\textcolor[rgb]{0,0,1}{0.0781}}
&\multicolumn{2}{|c|}{0.8060}&\multicolumn{2}{|c|}{\textcolor[rgb]{0,1,0}{0.0651}}
&\multicolumn{2}{|c|}{0.6845}&\multicolumn{2}{|c|}{0.1254}
&\multicolumn{2}{|c|}{0.8547}&\multicolumn{2}{|c|}{\textcolor[rgb]{0,0,1}{0.0685}}
&\multicolumn{2}{|c|}{0.7905}&\multicolumn{2}{|c|}{\textcolor[rgb]{0,1,0}{0.0709}}
\\
\multicolumn{4}{|c|}{V-A}
&\multicolumn{2}{|c|}{0.6128}&\multicolumn{2}{|c|}{0.1409}
&\multicolumn{2}{|c|}{0.8166}&\multicolumn{2}{|c|}{0.0811}
&\multicolumn{2}{|c|}{0.7346}&\multicolumn{2}{|c|}{0.0988}
&\multicolumn{2}{|c|}{0.6172}&\multicolumn{2}{|c|}{0.1367}
&\multicolumn{2}{|c|}{0.7641}&\multicolumn{2}{|c|}{0.1023}
&\multicolumn{2}{|c|}{0.6536}&\multicolumn{2}{|c|}{0.1044}
\\
\hline
\multicolumn{4}{|c|}{\textbf{DCL}~\cite{Li_2016_CVPR}}
&\multicolumn{2}{|c|}{\textcolor[rgb]{1,0,0}{0.6842}}&\multicolumn{2}{|c|}{0.1573}
&\multicolumn{2}{|c|}{0.8293}&\multicolumn{2}{|c|}{0.1495}
&\multicolumn{2}{|c|}{\textcolor[rgb]{1,0,0}{0.8533}}&\multicolumn{2}{|c|}{0.1359}
&\multicolumn{2}{|c|}{0.7141}&\multicolumn{2}{|c|}{0.1807}
&\multicolumn{2}{|c|}{0.8546}&\multicolumn{2}{|c|}{0.1513}
&\multicolumn{2}{|c|}{0.7946}&\multicolumn{2}{|c|}{0.1565}
\\
\multicolumn{4}{|c|}{\textbf{DS}~\cite{Li2016DeepSaliency}}
&\multicolumn{2}{|c|}{0.6028}&\multicolumn{2}{|c|}{0.1204}
&\multicolumn{2}{|c|}{0.8255}&\multicolumn{2}{|c|}{0.1216}
&\multicolumn{2}{|c|}{0.7851}&\multicolumn{2}{|c|}{0.0780}
&\multicolumn{2}{|c|}{0.6590}&\multicolumn{2}{|c|}{0.1760}
&\multicolumn{2}{|c|}{0.8445}&\multicolumn{2}{|c|}{0.0931}
&\multicolumn{2}{|c|}{0.7541}&\multicolumn{2}{|c|}{0.1233}
\\
\multicolumn{4}{|c|}{\textbf{ELD}~\cite{Lee_2016_CVPR}}
&\multicolumn{2}{|c|}{0.6109}&\multicolumn{2}{|c|}{\textcolor[rgb]{0,1,0}{0.0924}}
&\multicolumn{2}{|c|}{0.8102}&\multicolumn{2}{|c|}{0.0796}
&\multicolumn{2}{|c|}{0.7694}&\multicolumn{2}{|c|}{0.0741}
&\multicolumn{2}{|c|}{0.7180}&\multicolumn{2}{|c|}{\textcolor[rgb]{0,0,1}{0.1232}}
&\multicolumn{2}{|c|}{\textcolor[rgb]{1,0,0}{0.8715}}&\multicolumn{2}{|c|}{0.0670}
&\multicolumn{2}{|c|}{0.7591}&\multicolumn{2}{|c|}{0.1028}
\\
\multicolumn{4}{|c|}{\textbf{LEGS}~\cite{wang2015deep}}
&\multicolumn{2}{|c|}{0.5915}&\multicolumn{2}{|c|}{0.1334}
&\multicolumn{2}{|c|}{0.7853}&\multicolumn{2}{|c|}{0.1180}
&\multicolumn{2}{|c|}{0.7228}&\multicolumn{2}{|c|}{0.1193}
&\multicolumn{2}{|c|}{-}&\multicolumn{2}{|c|}{-}
&\multicolumn{2}{|c|}{0.8542}&\multicolumn{2}{|c|}{0.1034}
&\multicolumn{2}{|c|}{0.7358}&\multicolumn{2}{|c|}{0.1236}
\\
\multicolumn{4}{|c|}{\textbf{MDF}~\cite{zhao2015saliency}}
&\multicolumn{2}{|c|}{\textcolor[rgb]{0,1,0}{0.6442}}&\multicolumn{2}{|c|}{\textcolor[rgb]{1,0,0}{0.0916}}
&\multicolumn{2}{|c|}{0.8070}&\multicolumn{2}{|c|}{0.1049}
&\multicolumn{2}{|c|}{0.8006}&\multicolumn{2}{|c|}{0.0957}
&\multicolumn{2}{|c|}{0.7087}&\multicolumn{2}{|c|}{0.1458}
&\multicolumn{2}{|c|}{0.8419}&\multicolumn{2}{|c|}{0.0989}
&\multicolumn{2}{|c|}{\textcolor[rgb]{0,0,1}{0.8003}}&\multicolumn{2}{|c|}{0.1014}
\\
\multicolumn{4}{|c|}{\textbf{RFCN}~\cite{wang2016saliency}}
&\multicolumn{2}{|c|}{0.6265}&\multicolumn{2}{|c|}{\textcolor[rgb]{0,0,1}{0.1105}}
&\multicolumn{2}{|c|}{0.8340}&\multicolumn{2}{|c|}{0.1069}
&\multicolumn{2}{|c|}{\textcolor[rgb]{0,1,0}{0.8349}}&\multicolumn{2}{|c|}{0.0889}
&\multicolumn{2}{|c|}{\textcolor[rgb]{1,0,0}{0.7512}}&\multicolumn{2}{|c|}{0.1324}
&\multicolumn{2}{|c|}{0.8502}&\multicolumn{2}{|c|}{0.1166}
&\multicolumn{2}{|c|}{0.7667}&\multicolumn{2}{|c|}{0.1131}
\\
\hline
\multicolumn{4}{|c|}{\textbf{BL}~\cite{tong2015bootstrap}}
&\multicolumn{2}{|c|}{0.4988}&\multicolumn{2}{|c|}{0.2388}
&\multicolumn{2}{|c|}{0.6841}&\multicolumn{2}{|c|}{0.2159}
&\multicolumn{2}{|c|}{0.6597}&\multicolumn{2}{|c|}{0.2071}
&\multicolumn{2}{|c|}{0.5742}&\multicolumn{2}{|c|}{0.2487}
&\multicolumn{2}{|c|}{0.7675}&\multicolumn{2}{|c|}{0.1849}
&\multicolumn{2}{|c|}{0.7047}&\multicolumn{2}{|c|}{0.1856}
\\
\multicolumn{4}{|c|}{\textbf{BSCA}~\cite{qin2015saliency}}
&\multicolumn{2}{|c|}{0.5091}&\multicolumn{2}{|c|}{0.1902}
&\multicolumn{2}{|c|}{0.7048}&\multicolumn{2}{|c|}{0.1821}
&\multicolumn{2}{|c|}{0.6544}&\multicolumn{2}{|c|}{0.1748}
&\multicolumn{2}{|c|}{0.6006}&\multicolumn{2}{|c|}{0.2229}
&\multicolumn{2}{|c|}{0.8048}&\multicolumn{2}{|c|}{0.1535}
&\multicolumn{2}{|c|}{0.7062}&\multicolumn{2}{|c|}{0.1578}
\\
\multicolumn{4}{|c|}{\textbf{DRFI}~\cite{jiang2013salient}}
&\multicolumn{2}{|c|}{0.5504}&\multicolumn{2}{|c|}{0.1378}
&\multicolumn{2}{|c|}{0.7331}&\multicolumn{2}{|c|}{0.1642}
&\multicolumn{2}{|c|}{0.7218}&\multicolumn{2}{|c|}{0.1445}
&\multicolumn{2}{|c|}{0.6182}&\multicolumn{2}{|c|}{0.2065}
&\multicolumn{2}{|c|}{0.8068}&\multicolumn{2}{|c|}{0.1480}
&\multicolumn{2}{|c|}{0.7341}&\multicolumn{2}{|c|}{0.1334}
\\
\multicolumn{4}{|c|}{\textbf{DSR}~\cite{li2013saliency}}
&\multicolumn{2}{|c|}{0.5242}&\multicolumn{2}{|c|}{0.1389}
&\multicolumn{2}{|c|}{0.6621}&\multicolumn{2}{|c|}{0.1784}
&\multicolumn{2}{|c|}{0.6772}&\multicolumn{2}{|c|}{0.1422}
&\multicolumn{2}{|c|}{0.5575}&\multicolumn{2}{|c|}{0.2149}
&\multicolumn{2}{|c|}{0.7909}&\multicolumn{2}{|c|}{0.1579}
&\multicolumn{2}{|c|}{0.7116}&\multicolumn{2}{|c|}{0.1406}
\\
\hline
\end{tabular}
\vspace{0.7mm}
\caption{The F-measure and MAE of different saliency detection methods on five frequently used datasets.
The best three results are shown in \textcolor[rgb]{1,0,0}{red},~\textcolor[rgb]{0,1,0}{green} and \textcolor[rgb]{0,0,1}{blue}, respectively. The proposed methods rank first and second on these datasets.}
\label{table:fauc}
\end{center}
\vspace{-10mm}
\end{table*}
\subsection{Performance Comparison with State-of-the-art}
We compare the proposed UCF algorithm with other 10 state-of-the-art ones including 6
deep learning based algorithms (DCL~\cite{Li_2016_CVPR}, DS~\cite{Li2016DeepSaliency}, ELD~\cite{Lee_2016_CVPR}, LEGS~\cite{wang2015deep}, MDF~\cite{zhao2015saliency}, RFCN~\cite{wang2016saliency})
and 4 conventional counterparts (BL~\cite{tong2015bootstrap}, BSCA~\cite{qin2015saliency}, DRFI~\cite{jiang2013salient}, DSR~\cite{li2013saliency}).
The source codes with recommended parameters or the saliency maps of the competing methods are adopted for fair comparison.

As shown in Fig.~\ref{fig:PR-curve} and Tab.~\ref{table:fauc}, our proposed UCF model can consistently outperform
existing methods across almost all the datasets in terms of all evaluation metrics, which convincingly indicates the effectiveness of the proposed methods.
Refer to the supplemental material for more results on DUT-OMRON, HKU-IS, PASCAL-S and SOD datasets.

From these results, we have several fundamental observations:
(1) Our UCF model outperforms other algorithms on ECSSD and SED datasets with a large margin in terms of F-measure and MAE.
More specifically, our model improves the F-measure achieved by the best-performing existing algorithm by 3.9\% and 6.15\% on ECSSD and SED datasets, respectively.
The MAE is consistently improved.
(2) Although our proposed UCF is not the best on HKU-IS and PASCAL-S datasets, it is still very competitive (our model ranks the second on these datasets).
It is necessary to note that only the augmented MSRA10K dataset is used for training our model.
The RFCN, DS and DCL methods are pre-trained on the additional PASCAL VOC segmentation dataset~\cite{pascal-voc-2012}, which is overlaped with the PASCAL-S and HKU-IS datasets.
This fact may interpret their success on the two datasets.
However, their performance on other datasets is obviously inferior.
(3) Compared with other methods, our proposed UCF achieves lower MAE on most of datasets.
It means that our model is more convinced of the predicted regions by the uncertain feature learning.
%
%

\begin{figure*}
\centering
\begin{tabular}{@{}c@{}c@{}c@{}c@{}c@{}c@{}c@{}c@{}c@{}c}
\vspace{-1mm}
\includegraphics[width=0.09\linewidth,height=1.25cm]{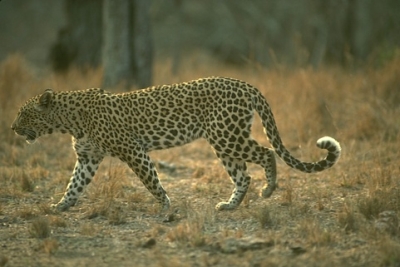} \ &
\includegraphics[width=0.09\linewidth,height=1.25cm]{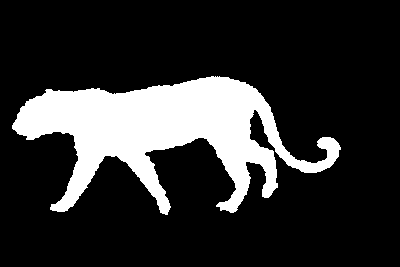} \ &
\includegraphics[width=0.09\linewidth,height=1.25cm]{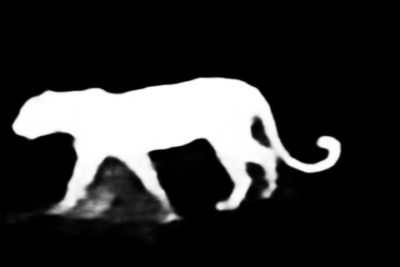} \ &
\includegraphics[width=0.09\linewidth,height=1.25cm]{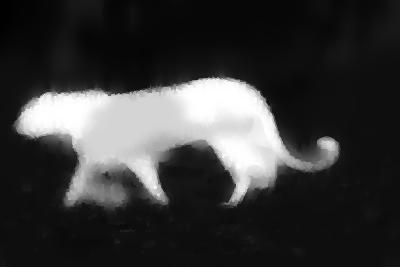} \ &
\includegraphics[width=0.09\linewidth,height=1.25cm]{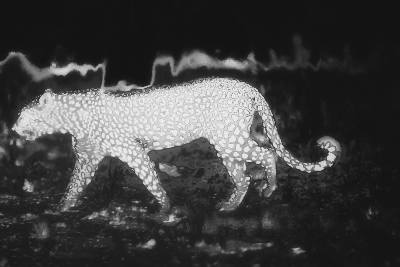} \ &
\includegraphics[width=0.09\linewidth,height=1.25cm]{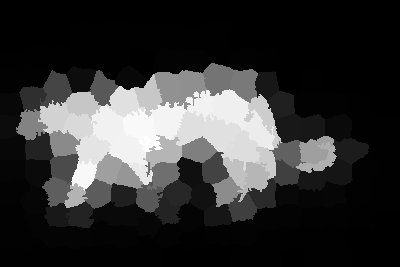} \ &
\includegraphics[width=0.09\linewidth,height=1.25cm]{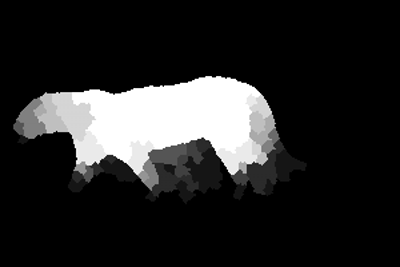} \ &
\includegraphics[width=0.09\linewidth,height=1.25cm]{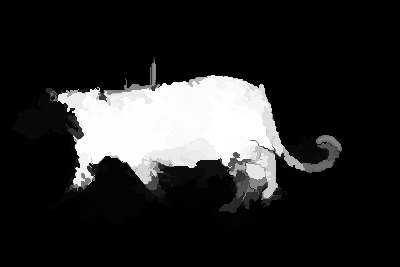} \ &
\includegraphics[width=0.09\linewidth,height=1.25cm]{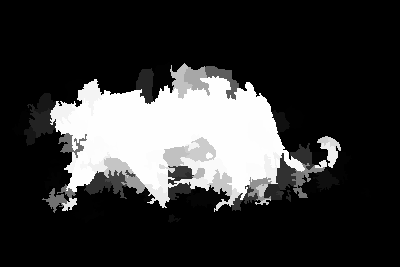} \ &
\includegraphics[width=0.09\linewidth,height=1.25cm]{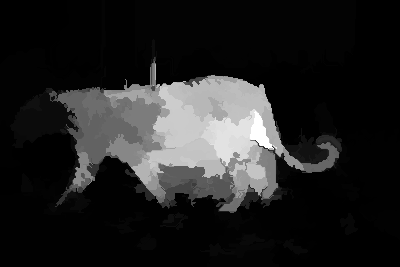} \ \\
\vspace{-1mm}
\includegraphics[width=0.09\linewidth,height=1.25cm]{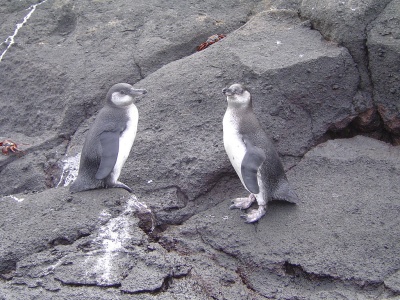} \ &
\includegraphics[width=0.09\linewidth,height=1.25cm]{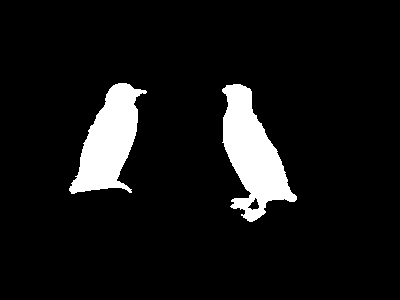} \ &
\includegraphics[width=0.09\linewidth,height=1.25cm]{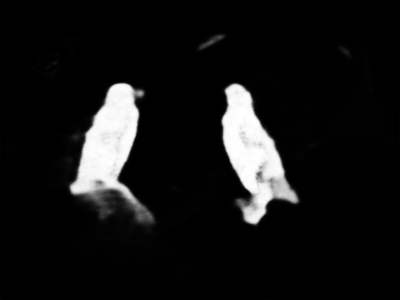} \ &
\includegraphics[width=0.09\linewidth,height=1.25cm]{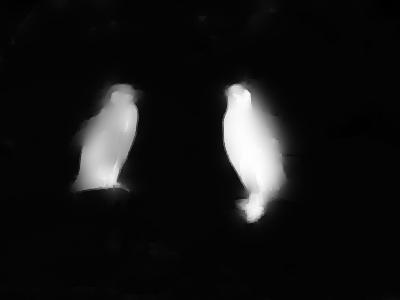} \ &
\includegraphics[width=0.09\linewidth,height=1.25cm]{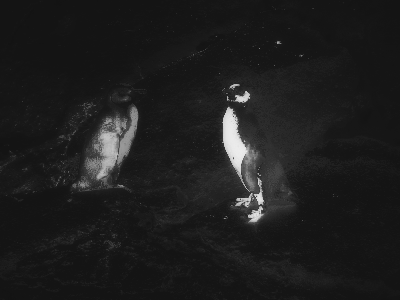} \ &
\includegraphics[width=0.09\linewidth,height=1.25cm]{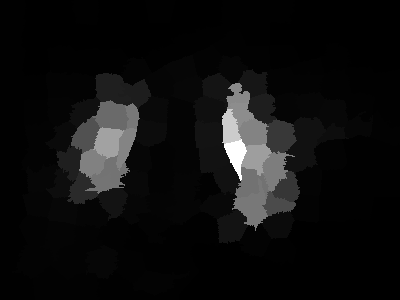} \ &
\includegraphics[width=0.09\linewidth,height=1.25cm]{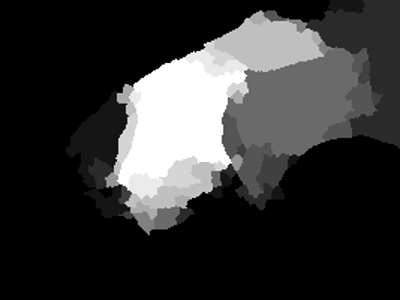} \ &
\includegraphics[width=0.09\linewidth,height=1.25cm]{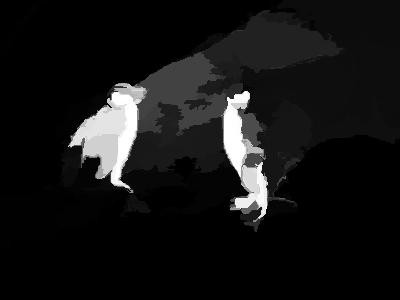} \ &
\includegraphics[width=0.09\linewidth,height=1.25cm]{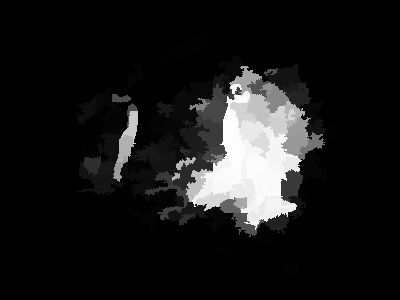} \ &
\includegraphics[width=0.09\linewidth,height=1.25cm]{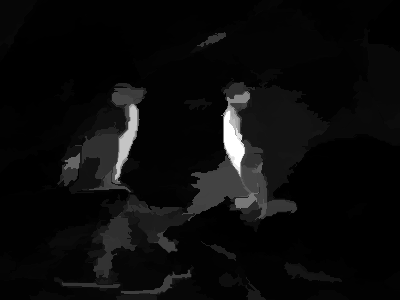} \ \\
\vspace{-1mm}
\includegraphics[width=0.09\linewidth,height=1.25cm]{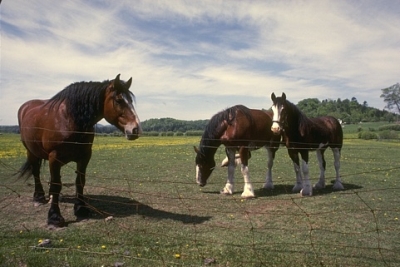} \ &
\includegraphics[width=0.09\linewidth,height=1.25cm]{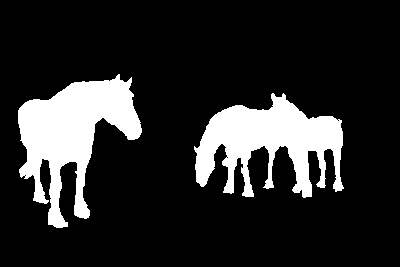} \ &
\includegraphics[width=0.09\linewidth,height=1.25cm]{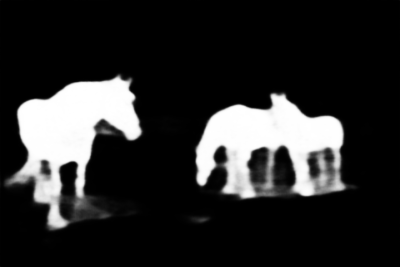} \ &
\includegraphics[width=0.09\linewidth,height=1.25cm]{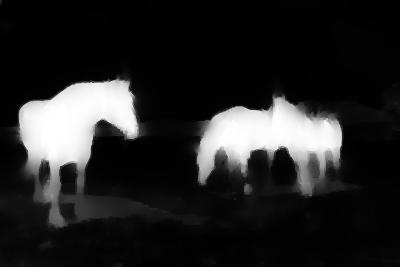} \ &
\includegraphics[width=0.09\linewidth,height=1.25cm]{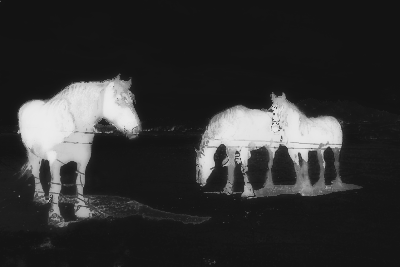} \ &
\includegraphics[width=0.09\linewidth,height=1.25cm]{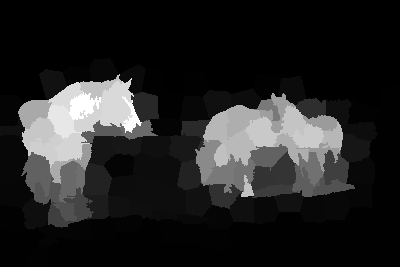} \ &
\includegraphics[width=0.09\linewidth,height=1.25cm]{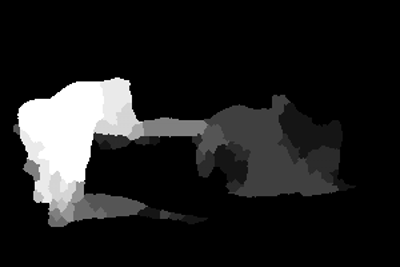} \ &
\includegraphics[width=0.09\linewidth,height=1.25cm]{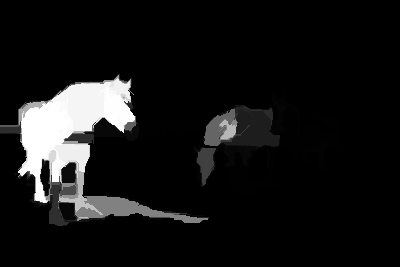} \ &
\includegraphics[width=0.09\linewidth,height=1.25cm]{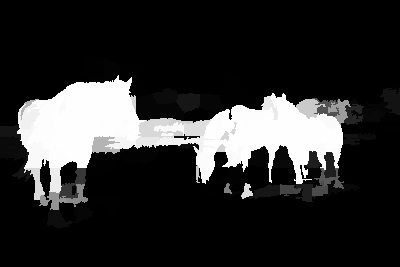} \ &
\includegraphics[width=0.09\linewidth,height=1.25cm]{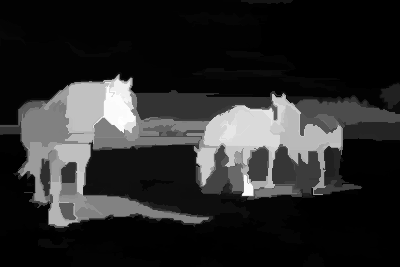} \ \\
\vspace{-1mm}
\includegraphics[width=0.09\linewidth,height=1.25cm]{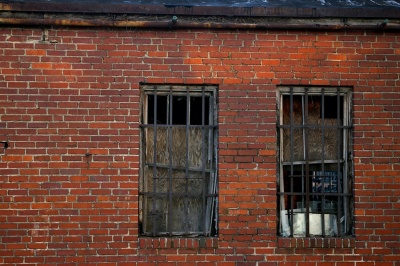} \ &
\includegraphics[width=0.09\linewidth,height=1.25cm]{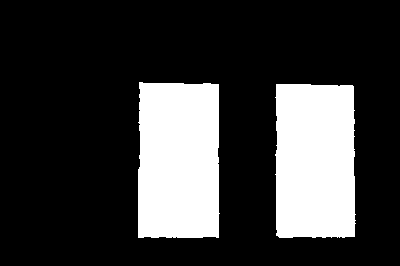} \ &
\includegraphics[width=0.09\linewidth,height=1.25cm]{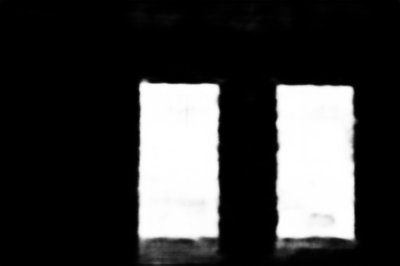} \ &
\includegraphics[width=0.09\linewidth,height=1.25cm]{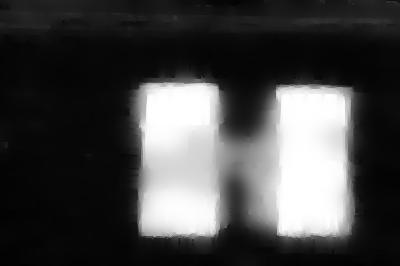} \ &
\includegraphics[width=0.09\linewidth,height=1.25cm]{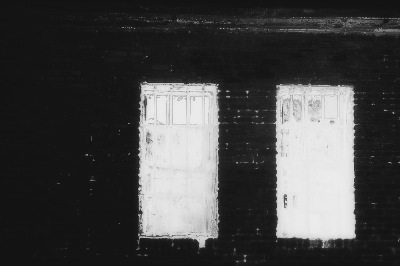} \ &
\includegraphics[width=0.09\linewidth,height=1.25cm]{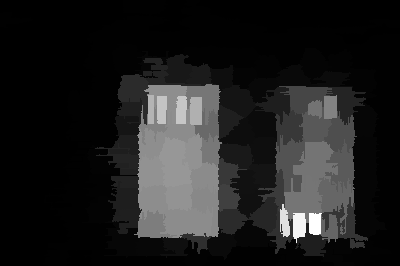} \ &
\includegraphics[width=0.09\linewidth,height=1.25cm]{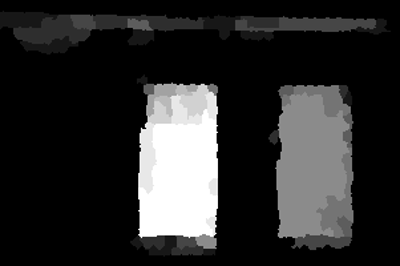} \ &
\includegraphics[width=0.09\linewidth,height=1.25cm]{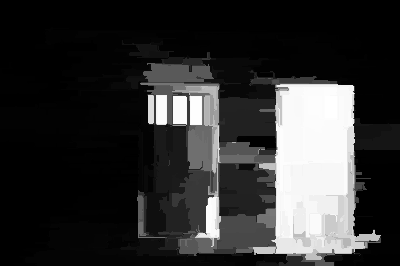} \ &
\includegraphics[width=0.09\linewidth,height=1.25cm]{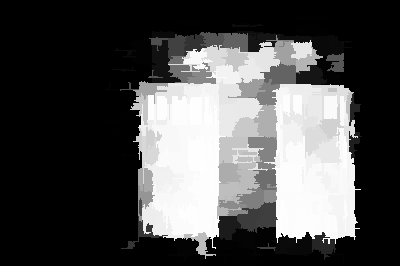} \ &
\includegraphics[width=0.09\linewidth,height=1.25cm]{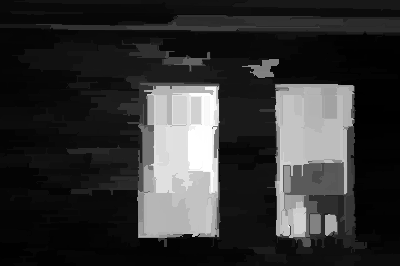} \ \\
\vspace{-1mm}
\includegraphics[width=0.09\linewidth,height=1.25cm]{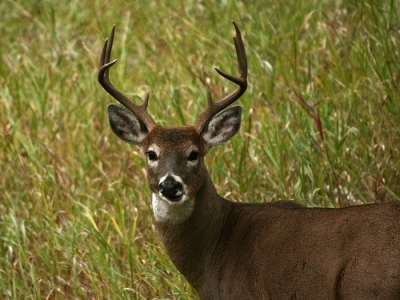} \ &
\includegraphics[width=0.09\linewidth,height=1.25cm]{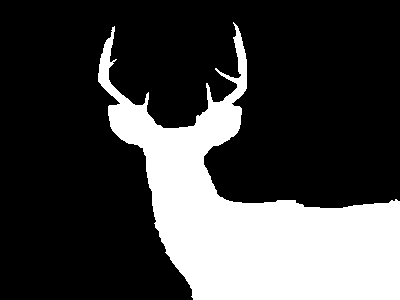} \ &
\includegraphics[width=0.09\linewidth,height=1.25cm]{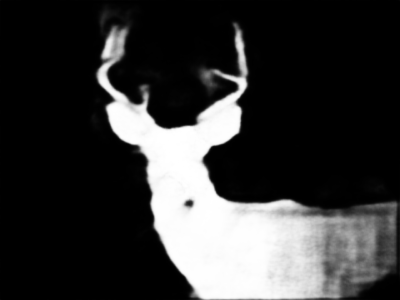} \ &
\includegraphics[width=0.09\linewidth,height=1.25cm]{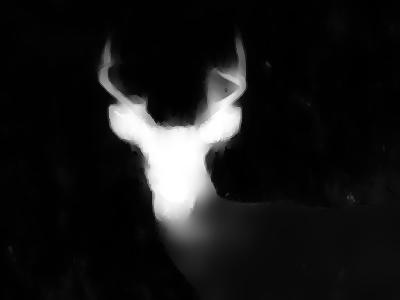} \ &
\includegraphics[width=0.09\linewidth,height=1.25cm]{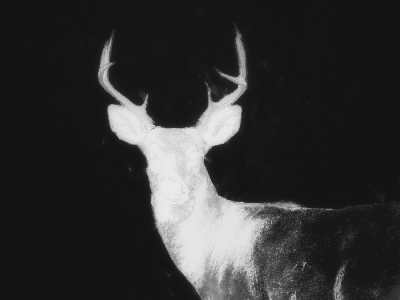} \ &
\includegraphics[width=0.09\linewidth,height=1.25cm]{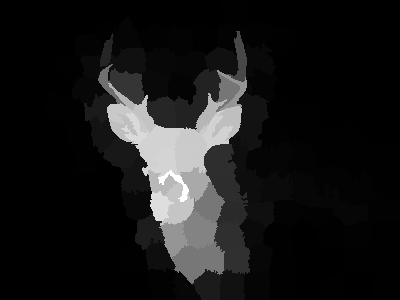} \ &
\includegraphics[width=0.09\linewidth,height=1.25cm]{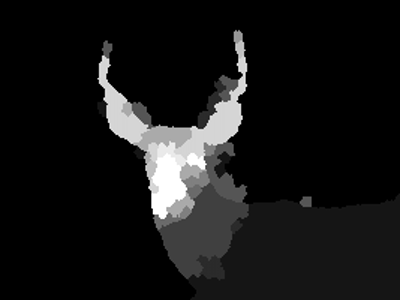} \ &
\includegraphics[width=0.09\linewidth,height=1.25cm]{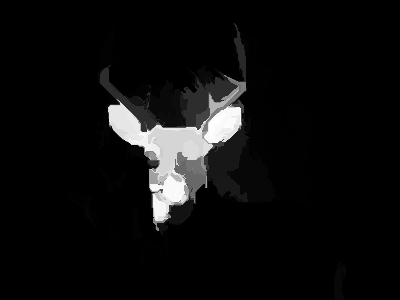} \ &
\includegraphics[width=0.09\linewidth,height=1.25cm]{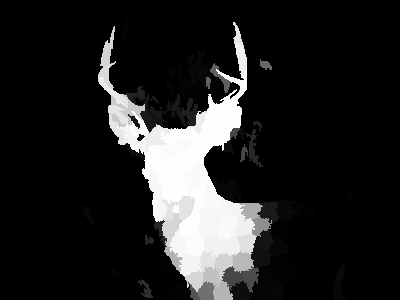} \ &
\includegraphics[width=0.09\linewidth,height=1.25cm]{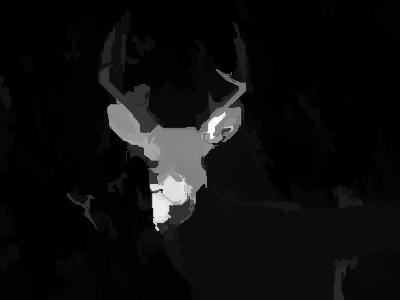} \ \\
\vspace{-1mm}
\includegraphics[width=0.09\linewidth,height=1.25cm]{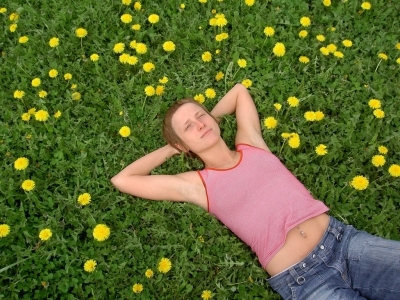} \ &
\includegraphics[width=0.09\linewidth,height=1.25cm]{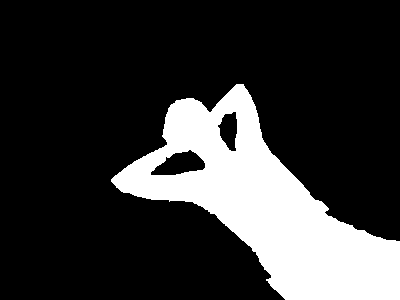} \ &
\includegraphics[width=0.09\linewidth,height=1.25cm]{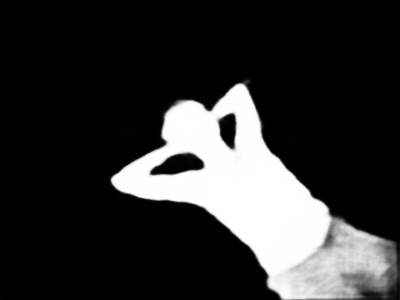} \ &
\includegraphics[width=0.09\linewidth,height=1.25cm]{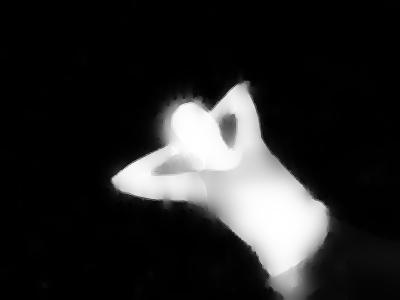} \ &
\includegraphics[width=0.09\linewidth,height=1.25cm]{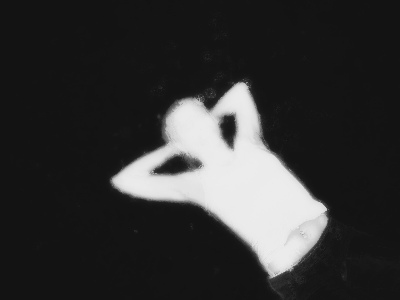} \ &
\includegraphics[width=0.09\linewidth,height=1.25cm]{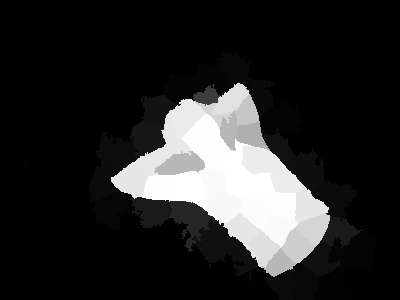} \ &
\includegraphics[width=0.09\linewidth,height=1.25cm]{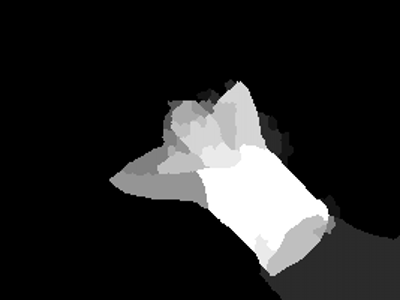} \ &
\includegraphics[width=0.09\linewidth,height=1.25cm]{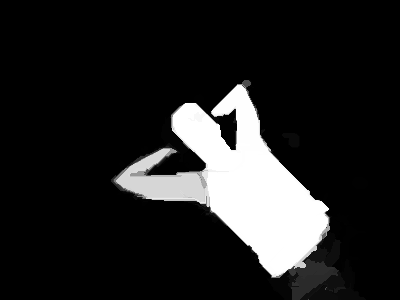} \ &
\includegraphics[width=0.09\linewidth,height=1.25cm]{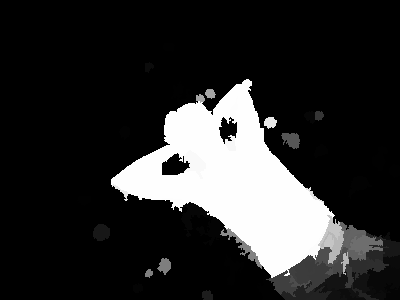} \ &
\includegraphics[width=0.09\linewidth,height=1.25cm]{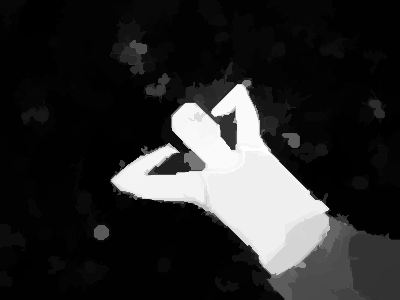} \ \\
\vspace{-1mm}
\includegraphics[width=0.09\linewidth,height=1.25cm]{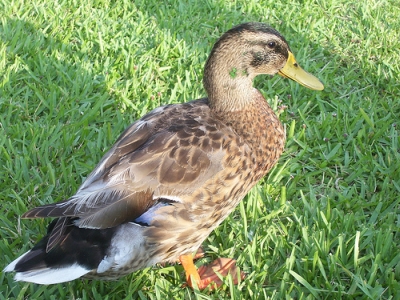} \ &
\includegraphics[width=0.09\linewidth,height=1.25cm]{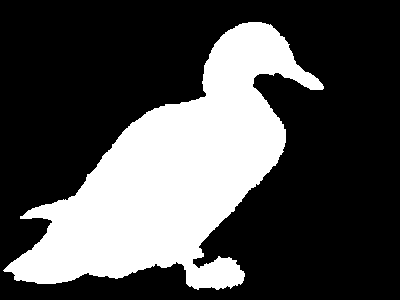} \ &
\includegraphics[width=0.09\linewidth,height=1.25cm]{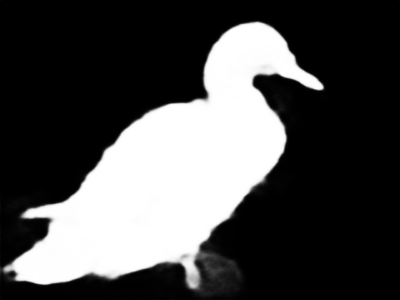} \ &
\includegraphics[width=0.09\linewidth,height=1.25cm]{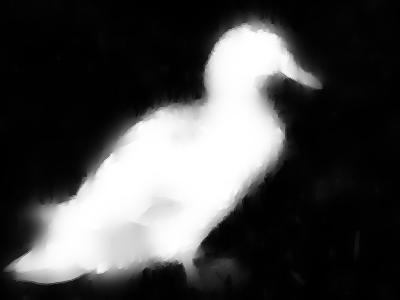} \ &
\includegraphics[width=0.09\linewidth,height=1.25cm]{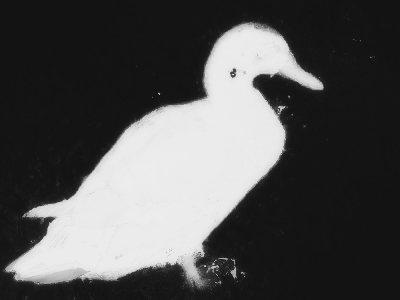} \ &
\includegraphics[width=0.09\linewidth,height=1.25cm]{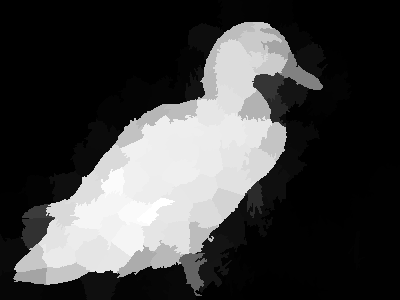} \ &
\includegraphics[width=0.09\linewidth,height=1.25cm]{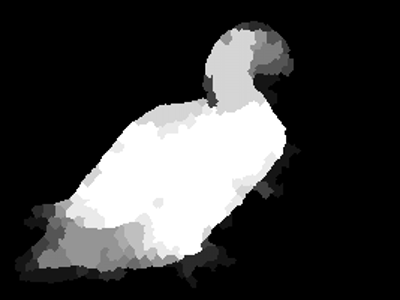} \ &
\includegraphics[width=0.09\linewidth,height=1.25cm]{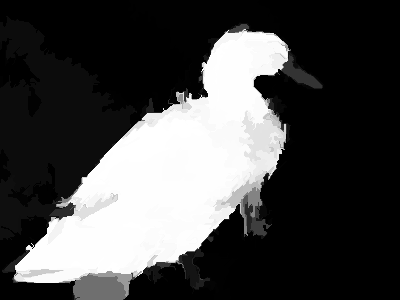} \ &
\includegraphics[width=0.09\linewidth,height=1.25cm]{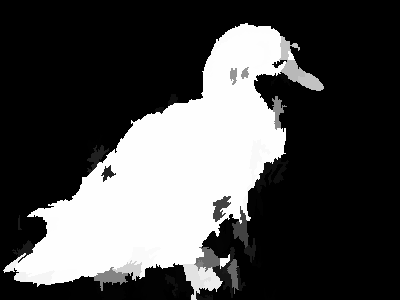} \ &
\includegraphics[width=0.09\linewidth,height=1.25cm]{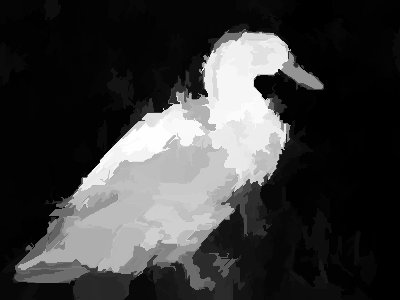} \ \\
\vspace{-1mm}
\includegraphics[width=0.09\linewidth,height=1.25cm]{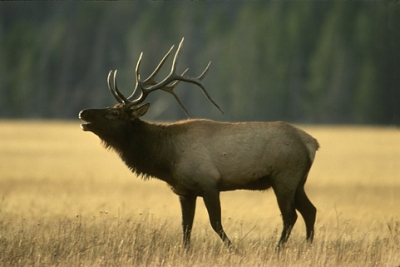} \ &
\includegraphics[width=0.09\linewidth,height=1.25cm]{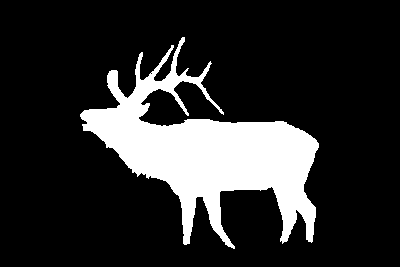} \ &
\includegraphics[width=0.09\linewidth,height=1.25cm]{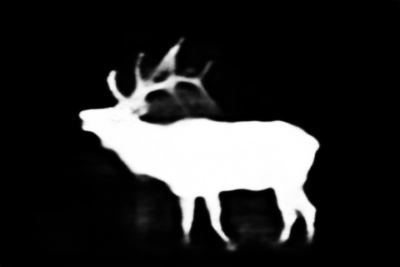} \ &
\includegraphics[width=0.09\linewidth,height=1.25cm]{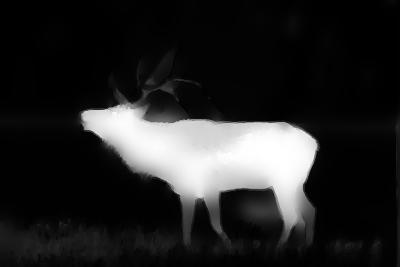} \ &
\includegraphics[width=0.09\linewidth,height=1.25cm]{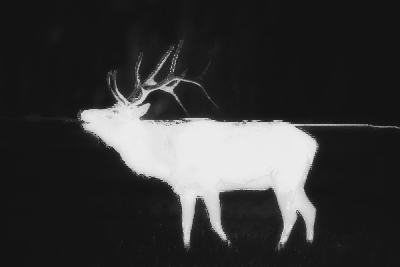} \ &
\includegraphics[width=0.09\linewidth,height=1.25cm]{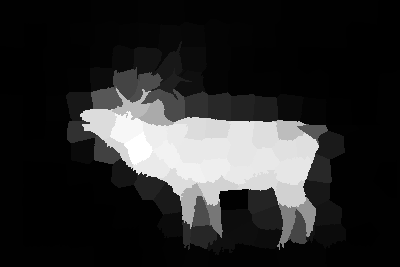} \ &
\includegraphics[width=0.09\linewidth,height=1.25cm]{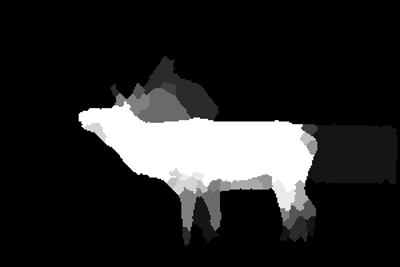} \ &
\includegraphics[width=0.09\linewidth,height=1.25cm]{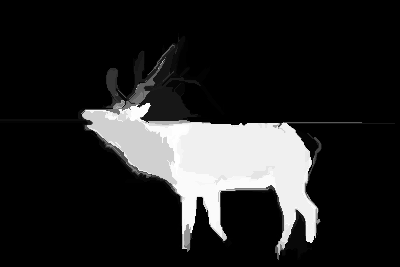} \ &
\includegraphics[width=0.09\linewidth,height=1.25cm]{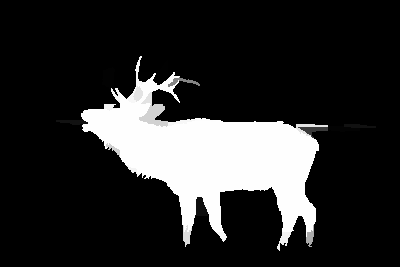} \ &
\includegraphics[width=0.09\linewidth,height=1.25cm]{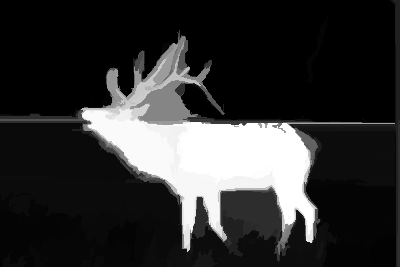} \ \\
{\small (a)} & {\small(b)} & {\small(c)} & {\small(d)} & {\small(e)}& {\small(f)}& {\small(g)}
& {\small(h)}& {\small(i)}& {\small(j)} \\
\end{tabular}
\caption{Comparison of saliency maps. (a) Input images; (b) Ground truth; (c) Our method; (d) RFCN; (e) DCL; (f) DS; (g) LEGS; (h) MDF; (i) ELD; (j) DRFI. More examples in each dataset can be found in the supplementary material.
\label{fig:map_examples}}
\vspace{-6mm}
\end{figure*}

The visual comparison of different methods on the typical images is shown in Fig.~\ref{fig:map_examples}.
Our saliency maps can reliably highlight the salient objects in various challenging scenarios, \eg, low contrast between objects and backgrounds (the first two rows), multiple disconnected salient objects (the 3-4 rows) and objects near the image boundary (the 5-6 rows). In addition, our saliency maps provide more accurate boundaries of salient objects (the 1, 3, 6-8 rows).
\setlength{\tabcolsep}{5pt}
\begin{table}
\begin{center}
\begin{tabular}{|c|c|c|c|c|c|c|}
\hline
Settings&V-A&V-B&V-C&V-D&V-E&UCF\\
\hline
+Dropout        &$\surd$&       &       &       &       &        \\
+R-Dropout      &       &$\surd$&$\surd$&       &       &$\surd$  \\
\hline
+Rest Deconv      &$\surd$&$\surd$&       &$\surd$&       &$\surd$    \\
+Inter          &       &       &$\surd$&       &$\surd$&$\surd$    \\
\hline
\end{tabular}
\end{center}
\vspace{-3mm}
\caption{Variants of our UCF model. Note that Rest Deconv and Inter indicate the hybrid upsampling method.}
\label{table:Ablation}
\vspace{-3mm}
\end{table}
\begin{figure}
\begin{tabular}{@{}c@{}c@{}c@{}c@{}c@{}}
\includegraphics[width=0.19\linewidth,height=1.2cm]{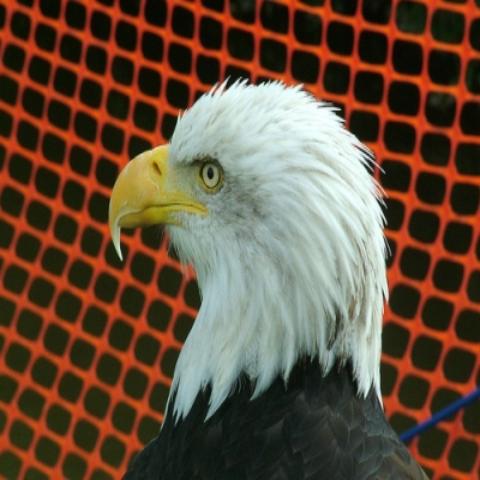} \ &
\includegraphics[width=0.19\linewidth,height=1.2cm]{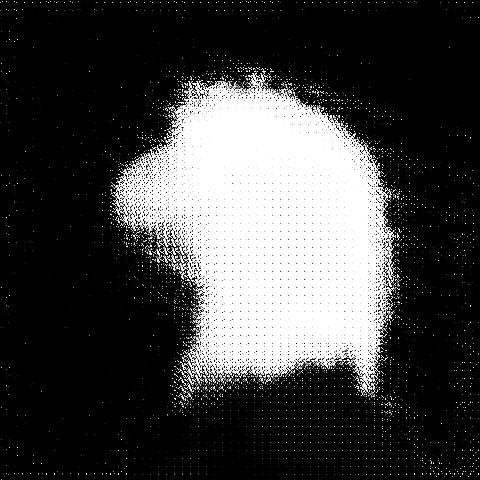} \ &
\includegraphics[width=0.19\linewidth,height=1.2cm]{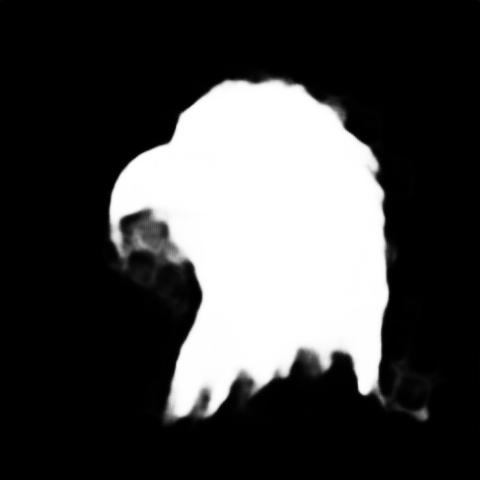} \ &
\includegraphics[width=0.19\linewidth,height=1.2cm]{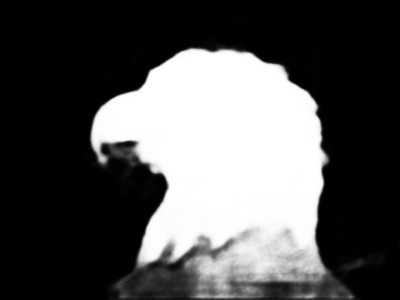} \ &
\includegraphics[width=0.19\linewidth,height=1.2cm]{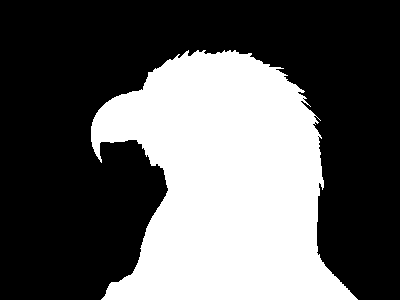} \ \\
{\small(a)} & {\small(b)} & {\small(c)} & {\small(d)} & {\small(e)}\\
\end{tabular}
\caption{Comparison of different upsampling algorithms. (a) Input image; (b) Deconvolution; (c) Interpolation; (d) Our method; (e) Ground truth. More examples in the supplementary material.
\label{fig:Ablation}}
\vspace{-6mm}
\end{figure}
{\flushleft\textbf{Ablation Studies:}} To verify the contributions of each component, we also evaluate several variants of the proposed UCF model with different settings as illustrated in Tab.~\ref{table:Ablation}.
The corresponding performance are reported in Tab.~\ref{table:fauc}.
The V-A model is an approximation of the DeconvNet~\cite{noh2015learning}.
The comparison between V-A and V-B demonstrates that our uncertain learning mechanism can indeed benefit to learn more robust features for accurate saliency inference.
The comparison between V-B and V-C shows the effects with two upsampling strategies.
Results imply that the interpolation strategy performs much better in saliency detection.
The joint comparison of V-B, V-C and UCF confirms that our hybrid upsampling method is capable of
better refining the output saliency maps.
An example on the visual effects is illustrated in Fig.~\ref{fig:Ablation}.
In addition, the V-D model and V-E model verify the usefulness of deconvolution and interpolation upsampling, respectively.
The V-B and V-C models achieve competitive even better results than other saliency methods. This further confirms the strength of our methods.
\vspace{-2mm}
\subsection{Generalization Evaluation}
To verify the generalization of our methods, we perform additional experiments on other pixel-wise vision tasks.
\vspace{-2mm}
{\flushleft\textbf{Semantic Segmentation:}} Following existing works~\cite{noh2015learning,kendall2015bayesian}, we simply change the classifier into 21 classes and perform the semantic segmentation task on the PASCAL VOC 2012 dataset~\cite{pascal-voc-2012}.
Our UCF model is trained with the PASCAL VOC 2011 training and validation data, using the Berekely's extended annotations ~\cite{BharathICCV2011}.
We achieve expressive results (mean IOU: 68.25, mean pix.accuracy: 92.19, pix.accuracy: 77.28), which are very comparable with other state-of-the-art segmentation methods.
In addition, though the segmentation performance gaps are not as large as in saliency detection, our new upsampling method indeed performs better than regular deconvolution (mean IOU: 67.45 vs 65.173, mean pix.accuracy: 91.21 vs 90.84, pix.accuracy: 76.18 vs 75.73).
\vspace{-2.45mm}
{\flushleft\textbf{Eye Fixation:}} The task of eye fixation prediction is essentially different from our classification task.
We use the Euclidean loss for the gaze prediction. We submit our results to servers of MIT300~\cite{Judd_2012}, iSUN~\cite{xu2015turkergaze} and SALICON~\cite{jiang2015salicon} benchmarks with standard setups.
Our model also achieves comparable results shown in Tab.~\ref{table:eye}.
All above results on semantic segmentation and eye fixation tasks indicate that our model has a strong generalization in other pixel-wise tasks.
\setlength{\tabcolsep}{4pt}
\begin{table}
\begin{center}
\begin{tabular}{|c|c|c|c|c|c|c|}
\hline
                         &AUC-J& sAUC &  CC  & NSS&  IG  \\
\hline
MIT300~\cite{Judd_2012}  &0.8584  &0.7109&0.7423&2.14&-    \\
\hline
iSUN~\cite{xu2015turkergaze}    &0.8615  &0.5498&0.8142&-   &0.1725\\
\hline
SALICON\cite{jiang2015salicon} &0.7621  &0.6326&0.8453&-   &0.3167\\
\hline
\end{tabular}
\end{center}
\vspace{-3mm}
\caption{Results on eye fixation datasets. Metrics in first row can be found in ~\cite{Judd_2012,jiang2015salicon}}
\label{table:eye}
\vspace{-6mm}
\end{table}
\vspace{-7mm}
\section{Conclusion}
\vspace{-1mm}
In this paper, we propose a novel fully convolutional network for saliency detection.
A reformulated dropout is utilized to facilitate probabilistic training and inference.
This uncertain learning mechanism enables our method to learn uncertain convolutional features and yield more accurate saliency prediction.
A new upsampling method is also proposed to reduce the artifacts of deconvolution operations, and explicitly enforce the network to learn accurate boundary for saliency detection.
Extensive evaluations demonstrate that our methods can significantly improve performance of saliency detection and show good generalization on other pixel-wise vision tasks.
%
\vspace{-3mm}
{\small {\flushleft\textbf{Acknowledgment}}.
We thank to Alex Kendall for sharing the SegNet code.
This paper is supported by the Natural Science Foundation of China \#61472060, \#61502070, \#61528101 and \#61632006.}
\vspace{-5mm}
{\small
\bibliographystyle{ieee}
\bibliography{egbib}

\begin{thebibliography}{10}\itemsep=-1pt

\bibitem{alexe2010object}
B.~Alexe, T.~Deselaers, and V.~Ferrari.
\newblock What is an object?
\newblock In {\em CVPR}, pages 73--80, 2010.

\bibitem{borj2015salient}
A.~Borji.
\newblock What is a salient object? a dataset and a baseline model for salient
  object detection.
\newblock {\em IEEE TIP}, 24(2):742--756, 2015.

\bibitem{borji2015salient}
A.~Borji, M.-M. Cheng, H.~Jiang, and J.~Li.
\newblock Salient object detection: A benchmark.
\newblock {\em IEEE TIP}, 24(12):5706--5722, 2015.

\bibitem{ChengPAMI}
M.-M. Cheng, N.~J. Mitra, X.~Huang, P.~H.~S. Torr, and S.-M. Hu.
\newblock Global contrast based salient region detection.
\newblock {\em IEEE TPAMI}, 37(3):569--582, 2015.

\bibitem{imagenet_cvpr09}
J.~Deng, W.~Dong, R.~Socher, L.-J. Li, K.~Li, and L.~Fei-Fei.
\newblock Imagenet:a large-scale hierarchical image database.
\newblock In {\em CVPR}, 2009.

\bibitem{ding2011importance}
Y.~Ding, J.~Xiao, and J.~Yu.
\newblock Importance filtering for image retargeting.
\newblock In {\em CVPR}, pages 89--96, 2011.

\bibitem{dosovitskiy2015inverting}
A.~Dosovitskiy and T.~Brox.
\newblock Inverting visual representations with convolutional networks.
\newblock {\em arXiv:1506.02753}, 2015.

\bibitem{Everingham2010ThePV}
M.~Everingham, L.~V. Gool, C.~K.~I. Williams, J.~M. Winn, and A.~Zisserman.
\newblock The pascal visual object classes (voc) challenge.
\newblock {\em IJCV}, 88:303--338, 2010.

\bibitem{pascal-voc-2012}
M.~Everingham, L.~Van~Gool, C.~K.~I. Williams, J.~Winn, and A.~Zisserman.
\newblock The {PASCAL} {V}isual {O}bject {C}lasses {C}hallenge 2012 {(VOC2012)}
  {R}esults.

\bibitem{gal2015dropout}
Y.~Gal and Z.~Ghahramani.
\newblock Dropout as a bayesian approximation: Insights and applications.
\newblock In {\em Deep Learning Workshop, ICML}, 2015.

\bibitem{BharathICCV2011}
B.~Hariharan, P.~Arbelaez, L.~Bourdev, S.~Maji, and J.~Malik.
\newblock Semantic contours from inverse detectors.
\newblock In {\em ICCV}, 2011.

\bibitem{he2015delving}
K.~He, X.~Zhang, S.~Ren, and J.~Sun.
\newblock Delving deep into rectifiers: Surpassing human-level performance on
  imagenet classification.
\newblock In {\em CVPR}, pages 1026--1034, 2015.

\bibitem{hinton2012improving}
G.~E. Hinton, N.~Srivastava, A.~Krizhevsky, I.~Sutskever, and R.~R.
  Salakhutdinov.
\newblock Improving neural networks by preventing co-adaptation of feature
  detectors.
\newblock {\em arXiv:1207.0580}, 2012.

\bibitem{hong2015online}
S.~Hong, T.~You, S.~Kwak, and B.~Han.
\newblock Online tracking by learning discriminative saliency map with
  convolutional neural network.
\newblock {\em arXiv:1502.06796}, 2015.

\bibitem{hou2007saliency}
X.~Hou and L.~Zhang.
\newblock Saliency detection: A spectral residual approach.
\newblock In {\em CVPR}, pages 1--8, 2007.

\bibitem{ioffe2015batch}
S.~Ioffe and C.~Szegedy.
\newblock Batch normalization: Accelerating deep network training by reducing
  internal covariate shift.
\newblock {\em arXiv:1502.03167}, 2015.

\bibitem{jiang2015salicon}
M.~Jiang, S.~Huang, J.~Duan, and Q.~Zhao.
\newblock Salicon: Saliency in context.
\newblock In {\em CVPR}, 2015.

\bibitem{jiang2013salient}
P.~Jiang, H.~Ling, J.~Yu, and J.~Peng.
\newblock Salient region detection by ufo: Uniqueness, focusness and
  objectness.
\newblock In {\em ICCV}, pages 1976--1983, 2013.

\bibitem{Judd_2012}
T.~Judd, F.~Durand, and A.~Torralba.
\newblock A benchmark of computational models of saliency to predict human
  fixations.
\newblock In {\em MIT Technical Report}, 2012.

\bibitem{kendall2015bayesian}
A.~Kendall, V.~Badrinarayanan, and R.~Cipolla.
\newblock Bayesian segnet: Model uncertainty in deep convolutional
  encoder-decoder architectures for scene understanding.
\newblock {\em arXiv:1511.02680}, 2015.

\bibitem{klein2011center}
D.~A. Klein and S.~Frintrop.
\newblock Center-surround divergence of feature statistics for salient object
  detection.
\newblock In {\em ICCV}, pages 2214--2219, 2011.

\bibitem{Lee_2016_CVPR}
G.~Lee, Y.-W. Tai, and J.~Kim.
\newblock Deep saliency with encoded low level distance map and high level
  features.
\newblock In {\em CVPR}, June 2016.

\bibitem{li2015visual}
G.~Li and Y.~Yu.
\newblock Visual saliency based on multiscale deep features.
\newblock In {\em CVPR}, pages 5455--5463, 2015.

\bibitem{Li_2016_CVPR}
G.~Li and Y.~Yu.
\newblock Deep contrast learning for salient object detection.
\newblock In {\em CVPR}, pages 478--487, 2016.

\bibitem{li2013saliency}
X.~Li, H.~Lu, L.~Zhang, X.~Ruan, and M.-H. Yang.
\newblock Saliency detection via dense and sparse reconstruction.
\newblock In {\em ICCV}, pages 2976--2983, 2013.

\bibitem{Li2016DeepSaliency}
X.~Li, L.~Zhao, L.~Wei, M.-H. Yang, F.~Wu, Y.~Zhuang, H.~Ling, and J.~Wang.
\newblock Deepsaliency: Multi-task deep neural network model for salient object
  detection.
\newblock {\em IEEE TIP}, 25(8):3919--3930, 2016.

\bibitem{li2014secrets}
Y.~Li, X.~Hou, C.~Koch, J.~Rehg, and A.~Yuille.
\newblock The secrets of salient object segmentation.
\newblock In {\em CVPR}, pages 280--287, 2014.

\bibitem{long2015fully}
J.~Long, E.~Shelhamer, and T.~Darrell.
\newblock Fully convolutional networks for semantic segmentation.
\newblock In {\em CVPR}, pages 3431--3440, 2015.

\bibitem{mahadevan2009saliency}
V.~Mahadevan and N.~Vasconcelos.
\newblock Saliency-based discriminant tracking.
\newblock In {\em CVPR}, pages 1007--1013, 2009.

\bibitem{marchesotti2009framework}
L.~Marchesotti, C.~Cifarelli, and G.~Csurka.
\newblock A framework for visual saliency detection with applications to image
  thumbnailing.
\newblock In {\em ICCV}, pages 2232--2239, 2009.

\bibitem{noh2015learning}
H.~Noh, S.~Hong, and B.~Han.
\newblock Learning deconvolution network for semantic segmentation.
\newblock In {\em ICCV}, pages 1520--1528, 2015.

\bibitem{odena2016deconvolution}
A.~Odena, V.~Dumoulin, and C.~Olah.
\newblock Deconvolution and checkerboard artifacts.
\newblock http://distill.pub/2016/deconv-checkerboard/, 2016.

\bibitem{qin2015saliency}
Y.~Qin, H.~Lu, Y.~Xu, and H.~Wang.
\newblock Saliency detection via cellular automata.
\newblock In {\em CVPR}, pages 110--119, 2015.

\bibitem{rother2004grabcut}
C.~Rother, V.~Kolmogorov, and A.~Blake.
\newblock Grabcut: Interactive foreground extraction using iterated graph cuts.
\newblock In {\em ACM TOG}, volume~23, pages 309--314, 2004.

\bibitem{salimans2016improved}
T.~Salimans, I.~Goodfellow, W.~Zaremba, V.~Cheung, A.~Radford, and X.~Chen.
\newblock Improved techniques for training gans.
\newblock {\em arXiv:1606.03498}, 2016.

\bibitem{Shi_2016_CVPR}
W.~Shi, J.~Caballero, F.~Huszar, J.~Totz, A.~P. Aitken, R.~Bishop, D.~Rueckert,
  and Z.~Wang.
\newblock Real-time single image and video super-resolution using an efficient
  sub-pixel convolutional neural network.
\newblock In {\em CVPR}, 2016.

\bibitem{siagian2007rapid}
C.~Siagian and L.~Itti.
\newblock Rapid biologically-inspired scene classification using features
  shared with visual attention.
\newblock {\em IEEE TPAMI}, 29(2):300--312, 2007.

\bibitem{simonyan2014very}
K.~Simonyan and A.~Zisserman.
\newblock Very deep convolutional networks for large-scale image recognition.
\newblock {\em arXiv:1409.1556}, 2014.

\bibitem{srivastava2014dropout}
N.~Srivastava, G.~E. Hinton, A.~Krizhevsky, I.~Sutskever, and R.~Salakhutdinov.
\newblock Dropout: a simple way to prevent neural networks from overfitting.
\newblock {\em JMLR}, 15(1):1929--1958, 2014.

\bibitem{sun2011scale}
J.~Sun and H.~Ling.
\newblock Scale and object aware image retargeting for thumbnail browsing.
\newblock In {\em ICCV}, pages 1511--1518, 2011.

\bibitem{tong2015bootstrap}
N.~Tong, H.~Lu, X.~Ruan, and M.-H. Yang.
\newblock Salient object detection via bootstrap learning.
\newblock In {\em CVPR}, pages 1884--1892, 2015.

\bibitem{vincent2016arithmetic}
D.~Vincent and V.~Francesco.
\newblock A guide to convolution arithmetic for deep learning.
\newblock {\em arXiv:1603.07285}, 2016.

\bibitem{vincent2010stacked}
P.~Vincent, H.~Larochelle, I.~Lajoie, Y.~Bengio, and P.-A. Manzagol.
\newblock Stacked denoising autoencoders: Learning useful representations in a
  deep network with a local denoising criterion.
\newblock {\em JMLR}, 11(Dec):3371--3408, 2010.

\bibitem{wang2015deep}
L.~Wang, H.~Lu, X.~Ruan, and M.-H. Yang.
\newblock Deep networks for saliency detection via local estimation and global
  search.
\newblock In {\em CVPR}, pages 3183--3192, 2015.

\bibitem{wang2016saliency}
L.~Wang, L.~Wang, H.~Lu, P.~Zhang, and X.~Ruan.
\newblock Saliency detection with recurrent fully convolutional networks.
\newblock In {\em ECCV}, pages 825--841, 2016.

\bibitem{wang2016kernelized}
T.~Wang, L.~Zhang, H.~Lu, C.~Sun, and J.~Qi.
\newblock Kernelized subspace ranking for saliency detection.
\newblock In {\em ECCV}, pages 450--466, 2016.

\bibitem{xu2015turkergaze}
P.~Xu, K.~A. Ehinger, Y.~Zhang, A.~Finkelstein, S.~R. Kulkarni, and J.~Xiao.
\newblock Turkergaze: Crowdsourcing saliency with webcam based eye tracking.
\newblock {\em arXiv preprint arXiv:1504.06755}, 2015.

\bibitem{yan2013hierarchical}
Q.~Yan, L.~Xu, J.~Shi, and J.~Jia.
\newblock Hierarchical saliency detection.
\newblock In {\em CVPR}, pages 1155--1162, 2013.

\bibitem{yang2013saliency}
C.~Yang, L.~Zhang, H.~Lu, X.~Ruan, and M.-H. Yang.
\newblock Saliency detection via graph-based manifold ranking.
\newblock In {\em CVPR}, pages 3166--3173, 2013.

\bibitem{zhao2015saliency}
R.~Zhao, W.~Ouyang, H.~Li, and X.~Wang.
\newblock Saliency detection by multi-context deep learning.
\newblock In {\em CVPR}, pages 1265--1274, 2015.

\end{thebibliography}
}
\end{document}